\documentclass[lettersize,journal]{IEEEtran}
\usepackage{graphicx}
\usepackage{cite}
\usepackage{picinpar}

\usepackage{url}
\usepackage{flushend}
\usepackage[latin1]{inputenc}
\usepackage{colortbl}
\usepackage{soul}
\usepackage{multirow}
\usepackage{pifont}
\usepackage{color}
\usepackage{alltt}
\usepackage[hidelinks]{hyperref}
\usepackage{enumerate}
\usepackage{siunitx}
\usepackage{breakurl}
\usepackage{epstopdf}
\usepackage{pbox}
\usepackage{subfigure}
\usepackage{color}
\usepackage{soul}
\usepackage{amsmath}
\usepackage{amsfonts,amssymb} 
\usepackage{newunicodechar} 
\usepackage{verbatim}
\usepackage{lettrine}

\usepackage{booktabs}
\usepackage{threeparttable}
\usepackage{array}
\usepackage{tabularx}

\usepackage{tabularx}

\usepackage{algorithm}
\usepackage{algpseudocode}
\algnewcommand\algorithmicevery{\textbf{every}}
\algdef{SE}[EVERY]{Every}{EndEvery}[1]{\algorithmicevery\ #1 \ \algorithmicdo}{\algorithmicend}

\soulregister\cite7 
\soulregister\citep7 
\soulregister\citet7 
\soulregister\ref7 
\soulregister\pageref7 
\definecolor{modify}{rgb}{0, 0, 0}
\definecolor{second}{rgb}{0, 0, 0}
\definecolor{third}{rgb}{0, 0, 0}

\title{A Shared Control Framework for Mobile Robots with Planning-Level Intention Prediction}

\author{Jinyu Zhang, Lijun Han, \emph{Member, IEEE}, Feng Jian, Lingxi Zhang and Hesheng Wang, \emph{Senior Member, IEEE}

\thanks{This work was supported by the Natural Science Foundation of China under Grant 62225309, U24A20278, 62361166632 and 62403311, in part by the Explorers Program of Shanghai (Basic Research Funding) under Grant 25TS1415100. \textit{(Corresponding Author: Hesheng Wang, Lijun Han.)}}

\thanks{J. Zhang, L. Han, and H. Wang are with School of Automation and Intelligent Sensing, Shanghai Jiao Tong university, Shanghai Key Laboratory of Navigation and Location Based Services, Shanghai 200240, China. (e-mail: jinyu.zhang@sjtu.edu.cn, lijun\_han@sjtu.edu.cn, wanghesheng@sjtu.edu.cn).}

\thanks{F. Jian is with the School of Electronic Information and Electrical Engineering, Shanghai Jiao Tong university, Shanghai 200240, China. (e-mail: azf-157@sjtu.edu.cn).}

\thanks{L. Zhang is with the School of Mechanical Science and Engineering, Huazhong University of Science and Technology, Wuhan 430074, China. (e-mail: lx\_zhang@hust.edu.cn).}

}

\newcommand{\nomrow}[2]{%
    \par\noindent
    \parbox[t]{0.48\columnwidth}{%
        \normalsize\raggedright\strut #1\strut%
    }%
    \hfill%
    \parbox[t]{0.52\columnwidth}{%
        \normalsize\raggedright\strut #2\strut%
    }%
    \par\addvspace{3pt}%
}

\begin{document}

\maketitle
\thispagestyle{empty}
\pagestyle{empty}

\begin{abstract}

\textcolor{modify}{In mobile robot shared control, effectively understanding human motion intention is critical for seamless human-robot collaboration. This paper presents a novel shared control framework with planning-level human intention prediction. It allows the operator to modify the robot's desired path based on inferred short-term motion intention. To represent future motion intentions, we introduce the concept of an intention domain and use it as a constraint for intention-aware path replanning. The intention-domain prediction and path replanning problems are jointly formulated as a Markov Decision Process and solved through deep reinforcement learning. In addition, a Voronoi-based human trajectory generation algorithm is further developed to train the policy entirely in simulation without human participation or demonstration data.  
\textcolor{third}{We validate our approach through simulation and real-world experiments. In the real-robot user study, we compared the proposed method with both direct human control and a heuristic shared-control baseline. The proposed method reduced human interaction time by 60.4\% and increased trajectory clearance by 13.2\% relative to direct human control, further reduced human interaction time by 10.1\% relative to the heuristic shared-control baseline, and achieved the lowest mean task completion time among the three control conditions.} Subjective evaluations also indicate improved usability and reduced perceived workload.}

\end{abstract}


{\color{modify}
\section*{NOMENCLATURE}
\nomrow{$\mathbf{p}=(x,y)\in\mathbb{R}^{2}$}{Robot center-of-mass coordinates}
\nomrow{$\theta\in\mathbb{R}$}{Angular difference of global and robot-local frames}
\nomrow{$\mathbf{u}=[v,\omega]\in\mathbb{R}^{2}$}{Control input in robot-local frame}
\nomrow{$\mathbf{u}^h=[v^h,\omega^h]\in\mathbb{R}^{2}$}{Human operator's control input}
\nomrow{$\mathbf{M},\tilde{\mathbf{M}}$}{Global map and egocentric map}
\nomrow{$\mathbf{g}_f,\tilde{\mathbf{g}}_f\in\mathbb{R}^{2}$}{Final goal in global and local frame}
\nomrow{$\{\mathbf{g}_s^1,\ldots,\mathbf{g}_s^n\}$}{\textcolor{second}{Set of sub-goals}}
\nomrow{\textcolor{second}{$\hat{\mathbf{g}}_s\in\mathbb{R}^{2}$}}{Predicted sub-goal}
\nomrow{\shortstack[l]{$\boldsymbol{\tau}^h=[\mathbf{p}_1^h,\ldots,\mathbf{p}_N^h] \in\mathbb{R}^{2N}$}}{Human short-term intention}
\nomrow{$\mathcal{D}_I,\tilde{\mathcal{D}}_I$}{Intention domain in global and local frame}
\nomrow{\textcolor{second}{\shortstack[l]{$\mathbf{x}_h,\mathbf{y}_h, \mathbf{p}_0^h\in\mathbb{R}^{2}$}}}{Basis vectors and origin of the local reference frame of the human}
\nomrow{$h,r,\phi$}{Geometric parameters of the intention domain}
\nomrow{$\boldsymbol{\gamma}^h,\tilde{\boldsymbol{\gamma}}^h$}{Historical path buffer}
\nomrow{$\boldsymbol{\tau}^r,\tilde{\boldsymbol{\tau}}^r$}{Robot desired path}
\nomrow{$T,\delta$}{High-level and low-level control cycles}
\nomrow{$W,H^r,H^h$}{Egocentric map width, observation length of robot desired path, historical buffer length}
\nomrow{$(\mathcal{S},\mathcal{A},\mathcal{R},f)$}{State/action spaces, reward, transition dynamics}
\nomrow{$s_t,a_t,r_t$}{State, action, reward at time step $t$}
\nomrow{$(a,\mathbf{x}_a)$}{Parameterized action}
\nomrow{$\pi_{\psi}$}{Policy parameterized by $\psi$}
\nomrow{\shortstack[l]{$R,r_{task}, r_{terminal},r_{reg}$}}{Total/task/terminal/regularization rewards}
}
	
\section{Introduction}

\subsection{Motivation}

\lettrine[lines=2]{M}{OBILE} robots have advanced significantly in locomotion, perception, and navigation. However, they still struggle to handle demanding real-world tasks such as search and rescue. Their limitations in perception and cognitive awareness prevent them from adapting to complex and unpredictable environments. A promising direction to overcome these challenges is the integration of a human operator into the system, which is often referred to as a shared control framework. It leverages the complementary strengths of humans and robots: human adaptability and contextual awareness on one side, and robotic efficiency and precision on the other. As a result, system performance can be substantially improved.

In many tasks, mobile robots are expected to reach a target location or follow a predefined path. However, in practical missions, unforeseen events such as sudden environmental changes make it difficult to define these goals or trajectories beforehand. Consequently, human operators must guide the robot in real time, drawing on their experience and situational awareness. For effective collaboration, the robot must also act intelligently, adjusting its behavior according to the ongoing task and environmental cues. Therefore, how to best coordinate the two sources of intelligence from the human and the robot to ensure both safety and efficiency during task execution becomes a challenge.

To solve this problem, a common approach is to have the human operator directly decide the motion of the robot and convey his/her intention explicitly. Following this setup, several shared control frameworks have been proposed to blend human and autonomous inputs; for example, to help avoid collisions \cite{jiangSharedControlKinematic2016a} or to provide assistive motion in a task-appropriate way \cite{gaoContextualTaskawareShared2014}. However, commanding direct motion inputs requires a high commitment of the human, especially in complex tasks or cluttered environments.

To reduce this burden, other approaches extend shared control to the planning level. In this setting, the robot generates an initial trajectory using prior information before task execution. During operation, the robot continuously predicts human motion intentions and adjusts its desired trajectory accordingly. Therefore, when the trajectory aligns with the operator's intention, the robot can move autonomously, effectively reducing the operator's workload. To implement this, \cite{loseyTrajectoryDeformationsPhysical2018} \cite{leiIntentionPredictionBased2022} rely on predefined models of user motion patterns and adjust the trajectory through deformation or replanning. However, due to the limited expressive capacity of explicit models, unmodeled human intentions may be ignored, thereby restricting robot control authority. To address this,\cite{ngItTakesTwo2023b} \cite{ngDiffusionCoPolicySynergistic2024} capture motion patterns from the human demonstration data and generate the predicted trajectory in a receding-horizon fashion. Although these methods can capture the inherent multimodality of human actions, i.e., the possibility of multiple highly distinct futures, the generated trajectories often lack environmental constraints, making it difficult to guarantee safe robot operation.


Motivated by the limitations of existing shared control paradigms, we propose a novel mobile robot shared control framework with planning-level human intention prediction. In this framework, the human operator can not only intervene in the robot's current motion through a local shared controller but also modify the robot's desired path to influence its future motion plan. We propose a path replanning algorithm to update the robot's desired path according to inferred human motion intentions, expressed as an intention domain. This domain is modeled as a conical region in two-dimensional space, encapsulating the likely area of the operator's intended movement over a short horizon. This representation enables the robot to infer short-term human goals while maintaining trajectory safety and efficiency. The proposed method integrates the reliability of rule-based planning with the flexibility of learning-based intention prediction. The algorithm performs replanning in two phases: in the first stage, it predicts the intention domain using contextual information and generates a trajectory segment within it. In the second stage, it completes the path by planning from the endpoint of the initial segment to the final task goal. We jointly model intention-domain prediction and path replanning as a Markov Decision Process, solved via deep reinforcement learning. A Voronoi-based trajectory generation algorithm is further developed to support learning entirely in simulation without requiring human participation or demonstration data.

\subsection{Related Work}

\subsubsection{Mobile Robot Shared Control}

Shared Control has played a major role in mobile robots as it has the potential to combine human strengths in global planning and coarse control with machine strengths in fine motion control \cite{udupaSharedAutonomyAssistive2023a}. It has been widely used in the field of teleoperation \cite{masoneSharedPlanningControl2018} 
\cite{elbouSmoothHumanRobot2025} \cite{liTrilateralSharedControl2025} \cite{coffeyCollaborativeTeleoperationHaptic2022}  \cite{coffeyReactiveSafeCoNavigation2023} \cite{liMitigatingOverAssistanceTeleoperated2025} \cite{rcar_shared_control} \cite{robot_learning_mobile_robot}, operation of personal mobility platforms \cite{leiIntentionPredictionBased2022} and co-transportation \cite{ngItTakesTwo2023b} \cite{ngDiffusionCoPolicySynergistic2024} \cite{maHumanRobotCollaboration2024} \cite{sirintunaObjectDeformationAgnosticFramework2024}. For assisting the teleoperation of a mobile robot, a shared control law based on a hysteresis switch was proposed in \cite{jiangSharedControlKinematic2016a} to guarantee the safety of the system without hitting any obstacles. To further reduce the operational and cognitive burden of humans, a bilateral shared framework was proposed in \cite{masoneSharedPlanningControl2018}, where the human operator was allowed to modify the shape of a planned path online. In \cite{elbouSmoothHumanRobot2025}, a shared controller for Uncrewed Ground Vehicles (UGV) teleoperation was introduced, effectively deploying human-robot shared control in the field. \cite{liTrilateralSharedControl2025} developed a novel trilateral shared control architecture based on the velocity and force coordination of a dual-user haptic tele-training system for a hexapod robot. \cite{coffeyCollaborativeTeleoperationHaptic2022} and \cite{coffeyReactiveSafeCoNavigation2023} have introduced a co-navigation algorithm, where the robot provides haptic guidance and path suggestions for both collision  avoidance and guidance to a target. 

\textcolor{modify}{Although these methods improve safety and reduce human effort, they do not explicitly predict human intention during task execution, which limits their ability to provide task-appropriate assistance to the human operator. A recent study \cite{chandraMultirobotnavigationsocial2026a} shows that mobile robot navigation often faces challenges caused by limited visibility and constrained spaces in common scenarios such as doorways, L-corners, and blind corners. In shared control settings, these scenario-level properties can also significantly increase the difficulty of human control. In such cases, by inferring the operator's motion intention and proactively adjusting its motion strategy, the robot can relieve the human operator from complex motion control and allow them to focus more on high-level task decision-making.}

\subsubsection{Human Intention Prediction}

In order to provide more useful assistance, researchers studying human-robot collaboration have dedicated significant effort to developing methods for human intention inference and action prediction. \cite{hoffmanInferringHumanIntent2024} divide relevant literature into three sections: inferring the human's intentions and goals, inferring specific collaborative features of the human, and predicting the human's future movement in space. The problem we focus on in this article can be considered as inferring the human's intentions and goals, that is, reasoning about the unobservable human goal-oriented constructs that underlie their behavior. Reference \cite{yowSharedAutonomyRobotic2024} formulates a shared autonomy framework for robotic manipulators as a discrete-action partially observable Markov decision process (POMDP). The system predicts human grasping intentions and selects among goal-directed actions, non-assistive actions, and information-gathering actions to support effective collaboration. Reference \cite{hanSharedControlPHRI2025} employs a fast randomized motion planner to generate multiple feasible motion plans as candidates for the predicted human intention and selects the best plan based on the energy function that includes human adjustment efforts. In \cite{luoHumanRobotShared2022a} , sub-goals are extracted from human demonstration trajectories as a representation of user intention in hydraulic manipulator teleoperation. Based on this, an intention prediction based arbitration rule was established to blend the controllers of the human and robot. \textcolor{modify}{Reference \cite{songRobotTrajectronTrajectory2024a} proposes a shared-control method that combines CVAE-based future trajectory prediction with artificial potential fields to provide intent-aware assistance in robot teleoperation tasks.} However, as we discussed earlier, above approaches reduce intentions to a selection of one of several goals or a limited number of strategies, ignoring those unmodeled intents. 

To address the limitations of explicit intention models, some researchers directly predict future human desired trajectory or actions in human-robot collaboration. The authors of \cite{mengHierarchicalHumanMotion2024} develop a Gaussian Process Regression (GPR)-based approach to predict future human motions in a human-robot collaborative sawing task. The approach in \cite{gaoHybridRecurrentNeural2023} introduces a hybrid recurrent neural network to infer human motion intentions in collaborative assembly tasks, thereby enhancing interaction efficiency and coordination. Reference \cite{franceschiDesignAssistiveController2025} proposed a learning-based model combining cascaded Long Short-Term Memory (LSTM) and Fully Connected (FC) layers to predict human desired trajectories in physical human-robot interaction. The approach in \cite{dominguez-vidalImprovingHumanRobotInteraction2023} predicts the human's force using a Deep Learning model in a Human-Robot collaborative object transportation task to improve human-robot interaction effectiveness. \textcolor{modify}{Reference \cite{zolotasDisentangledSequenceClustering2022} introduces a disentangled VAE-based clustering framework to separate time-invariant global intentions from time-varying local motion dynamics for human intention inference in robotic wheelchair navigation. These methods use sequential models to learn conditional distributions over time and infer future states by recursively applying transition functions. While technically capable of generating paths of any length, their uncertainty and computational costs increase substantially as the prediction horizon expands.}

\textcolor{modify}{In contrast, non-sequential models directly model the distribution over full trajectories without temporal factorization of the dynamics. The approach in \cite{katuwandeniyaExactlikelihoodUserIntention2022} introduces a Normalizing Flow-based generative model to estimate a distribution over future user-intended trajectories for shared-control navigation. The model permits trajectories to be assigned exact likelihoods, enabling the system to select the most likely future user intention from generated trajectory samples. Reference \cite{fangHumanRobotShared2026} predicts future human desired trajectories and forces with a diffusion model and designs an uncertainty-aware adaptive impedance control strategy for physical human-robot interaction. The authors of \cite{casadoNavigatingUncertaintyDiffusionBased2026} formulate wheelchair user intention estimation as a conditional trajectory generation problem and uses a diffusion model to capture the multimodality and uncertainty of human navigation behavior. However, to balance predictive accuracy and computational efficiency, these models typically restrict their prediction horizons. Moreover, in practice, they rely heavily on the quality of available training data for prediction accuracy and require an additional low-level controller to guarantee the safety of predicted actions.}

\textcolor{modify}{Large Language Models (LLMs) and Transformer-based methods have recently emerged as promising tools for human intention inference and shared control with their commonsense reasoning capabilities. Reference \cite{taoLAMSLLMDrivenAutomatic2025a} leverages an LLM to infer the user's intended control actions from task instructions, robot states, and object poses, enabling automatic mode switching for assistive robotic-arm teleoperation. The approach in\cite{liuCasperInferringDiverse2025} proposes CASPER, a VLM-based assistive teleoperation framework that infers diverse human intents from teleoperated user inputs and visual context, then executes the inferred skill through a parameterized skill library for long-horizon mobile manipulation tasks. Reference \cite{nguyenSAFeCopilotUnifiedShared2025a} proposes SAFe-Copilot, a VLM-based shared-autonomy framework that infers high-level driver intent from multimodal interventions and environmental context to arbitrate between human and autonomous driving plans. The authors of \cite{martinez-baselgaHeyRobotPersonalizing2025a} use LLMs and VLMs to infer user-specified navigation preferences from language or images and translate them into MPC cost terms, weights, and parameters for adaptive robot motion. Reference \cite{huangLITLargeLanguage2024} leverages LLMs and VLMs to construct a language-based probabilistic graphical model for tracking high-level human intentions over long-horizon human-robot collaboration tasks. However, many LLM-based methods require context information, such as robot state and task instructions, to be accurately expressed in natural language or grounded through visual-language perception before inference. Moreover, these approaches may rely on explicit verbal explanations or predefined skill libraries, whereas in shared control the operator's motion intention may be implicit in continuous control inputs rather than explicitly stated.}

\subsubsection{Human-in-the-Loop Reinforcement Learning}

Human-in-the-loop reinforcement learning is a topic of active research. Integrating human inputs into the learning process helps agents customize behavior and adapt to dynamic human needs, perceptions, and physical activities \cite{yauAugmentedIntelligencePerspective2024}. In many cases, it leverages human feedback to train autonomous agents, as a form of human-aided learning, to enhance the convergence rate and the quality of learned knowledge, e.g. the TAMER framework \cite{warnellDeepTAMERInteractive2018}. However, in shared control settings, the agent will always need to cooperate with the user to accomplish the task, due to the lack of information that is private to the user and relevant to the task. The authors of \cite{reddySharedAutonomyDeep2018a} propose a deep reinforcement learning (DRL) framework for model-free shared control. The success of assisting the user in drone-flying tasks illustrates the potential for DRL to enable flexible and practical assistive systems. In \cite{shaftiRealWorldHumanRobotCollaborative2020}, authors train a cooperative agent in a real-world collaborative maze game using DRL. The study results show the existence of human-robot co-learning and the potential to tailor RL policies to human personalization. To leverage the power of simulation to train RL policies in human-involved tasks, Iterative-Sim-to-Real (i-S2R) was proposed in \cite{abeyruwanISim2RealReinforcementLearning} . It bootstraps from a simple model of human behavior and alternates between training in simulation and deploying in the real world, refining both the human behavior model and the policy. Inspired by these approaches, our work jointly addresses intention-domain prediction and path replanning within a unified deep reinforcement learning framework. To accelerate the training process, we further introduce a human trajectory generation algorithm that replaces real human operation, enabling model training to be conducted entirely in simulation.

\subsection{Contributions}

In this work, we present a novel shared control framework for mobile robots that incorporates planning-level human intention prediction. This framework enables humans and robots to share authority at both control and planning levels in a flexible and safe manner.

The main contributions of this article are as follows.

\begin{itemize}
   
\item [1)] 
We introduce a closed-cone formulation of the intention domain and develop an intention-aware path replanning algorithm that leverages this representation to modify the robot's desired trajectory safely and efficiently.

  
\item [2)]
We jointly learn both intention-domain prediction and path replanning policies using deep reinforcement learning. Additionally, a Voronoi-based human trajectory generation method is proposed to enable fully simulation-based training without requiring human data.


\item [3)]
We conducted extensive experiments, including simulation studies on reward design, human-in-the-loop simulations, and a real-world user study with ten participants. The results demonstrate that our method reduces operator workload and enhances safety without compromising task efficiency compared with existing assistive shared control approaches.

\end{itemize}

\section{Problem Formulation and Algorithm Overview}

\subsection{Problem Formulation}

\begin{figure}
    \centering
    \includegraphics[width=8.5cm]{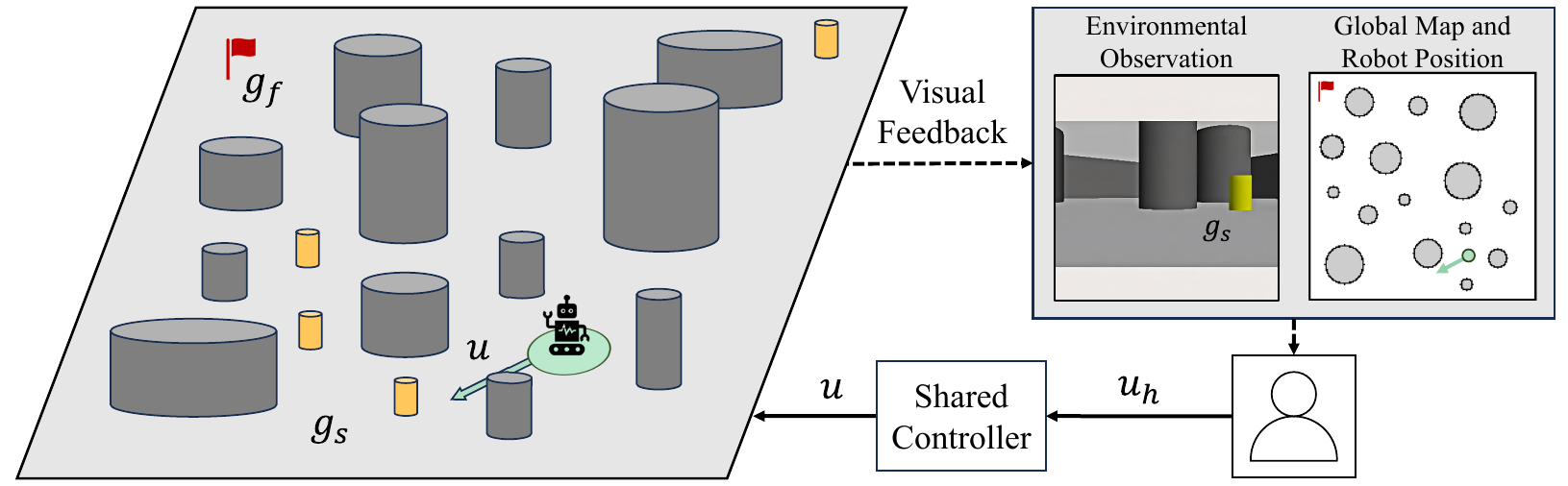}
    \caption{System configuration. \textcolor{modify}{The human operator guides the robot to visit all sub-goals $\mathbf{g}_s$ before reaching the final goal $\mathbf{g}_f$. The operator observes the environment and robot status through a monitor interface. Based on the visual feedback, the operator provides control commands $\mathbf{u}^h$ to the shared controller, which computes the robot control input $\mathbf{u}$ for robot motion control.}}
    \label{task_config}
\end{figure}

The configuration of the system is shown in Fig. \ref{task_config}, where a human operator shares control with an autonomous mobile robot. The robot dynamics are defined as:
\begin{equation}\label{dynamics}
\begin{aligned}
\begin{bmatrix}
\dot{x} \\
\dot{y} \\
\dot{\theta}
\end{bmatrix}
=
\begin{bmatrix}
\cos\theta & 0 \\
\sin\theta & 0 \\
0 & 1
\end{bmatrix}
\begin{bmatrix}
v \\
\omega
\end{bmatrix},
\end{aligned}
\end{equation}
where $\mathbf{p} = (x, y)$ denotes the Cartesian coordinates of the center of mass of the robot in the global reference frame and $\theta$ denotes the angular difference between the global and local reference frames. The control input $\mathbf{u}=[v, \omega]$ consists of the forward velocity $v$ and angular velocity $\omega$, defined in the robot's local frame.

Similar to autonomous navigation, the task goal includes a final goal $\mathbf{g}_f$, such as a supply station in a search-and-rescue task. Beyond this, the task involves a set of sub-goals $\{\mathbf{g}^1_s, \dots, \mathbf{g}^n_s\}$, which emerge during execution (e.g., victims detected during the mission). The human operator identifies these sub-goals through real-time environmental observation, such as the onboard camera view, and provides control inputs $\mathbf{u}^h = [v^h, \omega^h]$ via an interface (e.g., joystick) to guide the robot toward them.

We assume that the robot has a prior map of the environment, represented as an occupancy grid, and can localize with respect to this map. Additionally, to support the operator's situational awareness, both the global map and the robot's current position are made available to the operator.

\textit{Problem definition}: Given the system dynamics, a global map $\mathbf{M}$, a final goal $\mathbf{g}_f$, and a set of sub-goals $\{ \mathbf{g}^1_s, \dots, \mathbf{g}^n_s \}$, the human operator is required to guide the robot through all sub-goals before reaching the final goal. The operator has access to the global map, as well as the positions of the robot and the final goal, but the sub-goal locations are unknown in advance. Instead, the operator identifies sub-goals from environmental observations and provides control inputs $\mathbf{u}^h = [v^h, \omega^h]$ to direct the robot toward them. The objective is to compute the robot's control input $\mathbf{u}$ such that task efficiency and safety are maintained while minimizing the operator's workload.


\subsection{Algorithm Overview}

\begin{figure*}
    \centering
    \includegraphics[width=15.5cm]{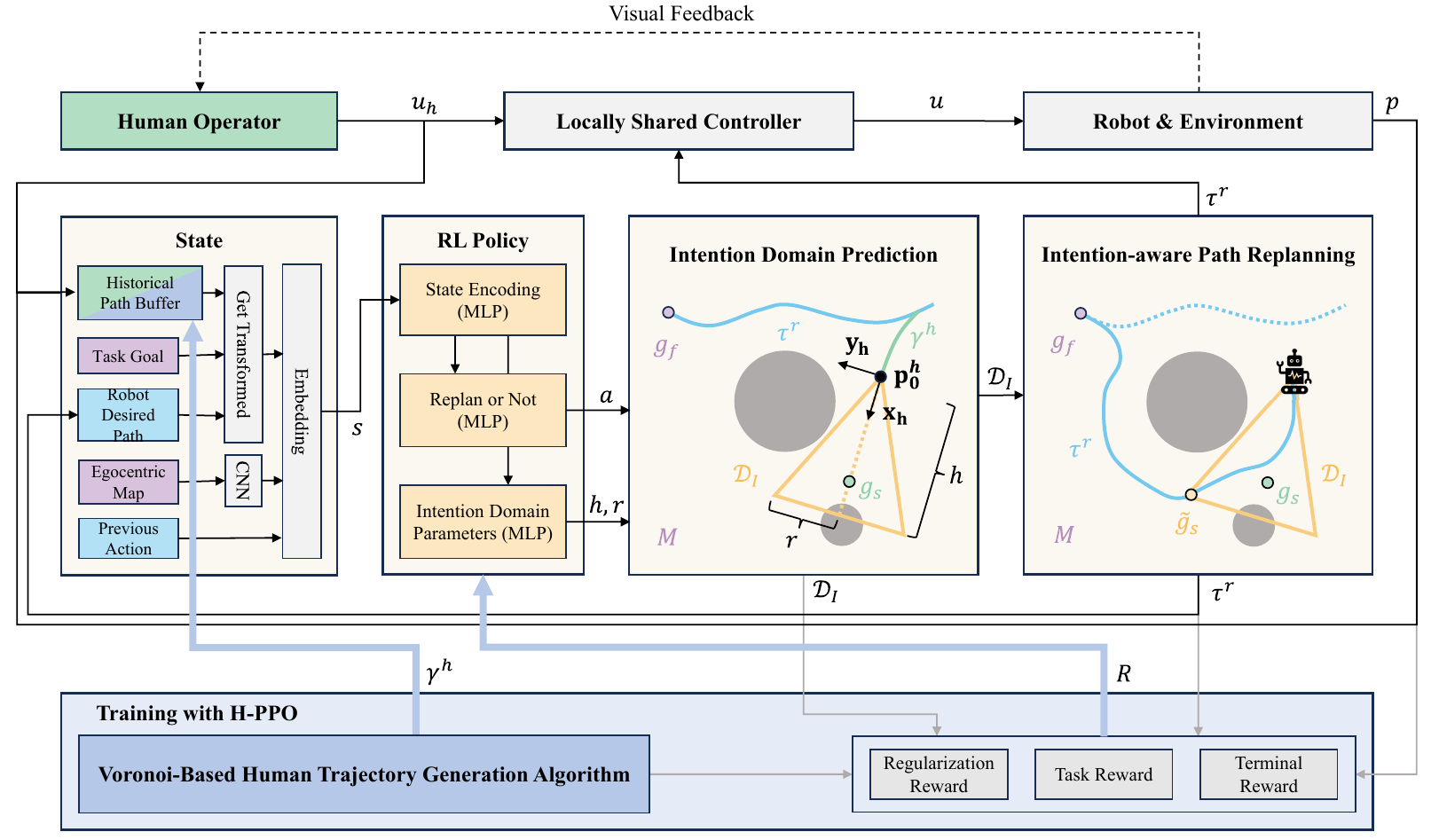}
    \caption{The proposed shared control framework. 
    \textcolor{modify}{The locally shared controller computes safe robot control commands $\mathbf{u}$ based on the human control input $\mathbf{u}^h$ and the robot's desired path $\tau^r$. when the human operator intervenes, the reinforcement learning policy jointly encodes the environment, task priors, and the motion intentions of both the robot and the human, and determines whether to trigger path replanning and, if replanning is triggered, predicts the geometric parameters $(h,r)$ of the intention domain $\mathcal{D}_I$. During replanning, the predicted intention domain is constructed and then used as a constraint to generate a new robot desired path $\tau^r$ that is consistent with the inferred human intention. The reinforcement learning policy is trained in simulation using H-PPO, where the human desired trajectories are generated by a Voronoi-based human trajectory generation algorithm without requiring human data.}
    Purple denotes environment and task priors, sky blue indicates robot intentions, green represents human motion intentions, and yellow shows inferred intentions in the form of an intention domain. Dark blue components correspond to modules used for policy training.}
    \label{framework}
\end{figure*}

The proposed shared control framework is illustrated in Fig. \ref{framework}.

At the beginning of execution, the robot plans an initial path based on prior task knowledge and the global map, navigating autonomously toward the common goal. The current pose and planned path of the robot, global map, and onboard camera observations are presented visually to the operator. 

When the current plan aligns with the operator's intent, the robot proceeds autonomously, thereby reducing the operational effort of the operator and allowing the operator to concentrate on higher-level situational awareness. However, when the operator intends to redirect the robot toward a sub-goal, control commands $\mathbf{u}^h$ are issued through an interface. The shared controller integrates this command with contextual information to update system inputs $\mathbf{u}$. At the same time, the planned path of the robot is adjusted accordingly. Once the revised path approaches the sub-goal, as judged by the operator, the operator can return control authority to the robot.

To realize this interaction, we introduce a path replanning framework with intention domain prediction, enabling real-time and intention-aware path adaptation. Within this framework, a reinforcement learning (RL) agent is trained to determine the triggering policy for path replanning and, if triggered, to infer the corresponding intention domain. Furthermore, a shared control module implements a switching strategy: When the operator provides input, a modified shared dynamic window approach (DWA) controller generates safe control commands that align with human intent under kinematic and obstacle constraints; otherwise, a path-tracking controller computes inputs to keep the robot on its planned trajectory.

\section{Path Replanning Algorithm with Intention Domain Prediction}

\subsection{Intention Domain}\label{ID_sec}

\textcolor{modify}{We categorize human intention into long-term and short-term components. The long-term intention is defined as reaching the final goal safely and efficiently. The path to the final goal planned by the robot prior to task execution aligns with the human long-term intention. However, during task execution, additional sub-goals may emerge dynamically and may not lie near the initially planned path. To search for or mark these sub-goals, the human operator may deviate from the original path. We interpret this deviation as the short-term intention. If this short-term intention can be predicted in a timely and accurate way, the robot can replan its desired path accordingly, enabling the human to hand back control authority to the robot and reducing the operational burden. }

\textcolor{modify}{Short-term human intention is represented as a sequence of equally spaced points,}
\begin{equation}\label{human_intention}
\boldsymbol{\tau}^h = [\mathbf{p}^h_1, \: \mathbf{p}^h_2, \: \mathbf{p}^h_3, \:\dots, \mathbf{p}^h_{N}] \in \mathbb{R}^{2N},
\end{equation}
where $N$ is the number of points and the spacing $\Delta d$ is determined by the grid map resolution. As we assume that the robot tracks these points at maximum admissible velocity, higher-order motion information (e.g., velocity, acceleration) is excluded from this representation. \textcolor{modify}{Unless otherwise specified, ``human intention" in the remainder of this paper refers to short-term intention.}

\textcolor{modify}{We adopt the paradigm of planning-based methods for human motion prediction \cite{rudenkoHumanMotionTrajectory2020}. These approaches model humans as rational agents, explicitly reasoning about their goals and computing policies to reach them.} Similar to the rational human, the mobile robot in shared control also needs to generate an optimal path (e.g., minimizing distance or maximizing smoothness) to lead the operator toward the intended goal. However, planning-based prediction assumes predefined goals and generates optimal paths by reasoning over these goals. \textcolor{modify}{This assumption does not hold in shared control settings where sub-goals emerge dynamically.} 

To address this limitation, we introduce the concept of the intention domain. \textcolor{modify}{It is defined as a non-empty, connected, and closed set in $\mathbb{R}^2$ that contains the short-term human intention.} Intention prediction is thus decomposed into two steps: constraining the planning space to the predicted domain and generating a path within it using planning-based reasoning.

\textcolor{modify}{We define the intention domain in the local reference frame of the human. Unlike the robot's local reference frame, which is determined by the robot's position and orientation, the local reference frame of the human is determined by the historical path buffer $\boldsymbol{\gamma}^h = \{\mathbf{p}_{-H^h}^h, \dots, \mathbf{p}_{-1}^h, \mathbf{p}_0^h\}$, where $H^h$ is the buffer length. It stores the robot's motion history during human intervention. Each time the robot travels more than $\Delta d$ under human intervention, its current position is added to the buffer. When the robot moves autonomously more than $\Delta d$, the buffer is cleared. We choose the latest point $\mathbf{p}_0^h$ in the buffer as the origin point. The orthonormal basis $\{\mathbf{x}_{h}, \mathbf{y}_{h}\}$ defines two axes relative to $\mathbf{p}_0^h$. The direction of $\mathbf{x}_h$ is derived from a sliding-window average of historical motion in the buffer $\boldsymbol{\gamma}^h$, as:}
\begin{equation}
\label{IntentD2}
\begin{array}{l}
\textcolor{modify}{\mathbf{x}_h = \frac{\sum_{i=1}^{H^h}(\mathbf{p}_{1 - i}^h - \mathbf{p}_{- i}^h)}{||\sum_{i=1}^{H^h}(\mathbf{p}_{1 - i}^h - \mathbf{p}_{- i}^h)||}.}
\end{array}
\end{equation}

\textcolor{modify}{In Section \ref{sec_GeneraID}, we compare different choices of intention domain shape. Here we introduce the most effective one. Let $\mathbf{q} = [x, y]^T \in \mathbb{R}^2$ denote the coordinates of an arbitrary point within the local reference frame of the human. As shown in Fig. \ref{framework}, the local intention domain $\tilde{\mathcal{D}}_I$ is defined as a closed (truncated) cone oriented along the positive $\mathbf{x}_h$ axis:
\begin{equation}
\label{IntentD1}
\begin{array}{l}
\tilde{\mathcal{D}}_I(h, r) = \left\{ \mathbf{q} \in \mathbb{R}^2 \:\: \bigg| \:\: x \in [0, h], \:\: |y| \le \frac{r}{h}x \right\},
\end{array}
\end{equation}
where the geometric parameters $h$ and $r$ represent the height and top half-width (radius) of the cone, respectively. The half-opening angle is naturally defined as $\phi = \arctan(\frac{r}{h}) \in [0, \frac{\pi}{2})$.}

\textcolor{modify}{The height $h$ determines the prediction horizon of the short-term intention, while the radius $r$ constrains the alignment between the predicted path and the operator's current heading. For a fixed opening angle, the domain expands nonlinearly with distance, reflecting the increasing uncertainty of human intention over time. To utilize the intention domain for motion planning or obstacle evaluation, the locally defined shape $\tilde{\mathcal{D}}_I$ is mapped back to the global coordinate space.} 

\textcolor{modify}{The intention domain's geometric parameters, height $h$ and radius $r$, are estimated from contextual information such as obstacle distribution and goal location.} In the following sections, we integrate this prediction process with path replanning, formulating the problem as a Markov Decision Process and solving it via deep reinforcement learning.

\subsection{Intention-aware Path Replanning}
{\color{modify}
To generate a safe path that aligns with both short-term and long-term intentions, the proposed intention-aware path replanning algorithm consists of two stages. In the first stage, an intention domain is inferred from contextual cues, and a path segment $\boldsymbol{\tau}^{rs}$ constrained within this domain is generated to capture the short-term intention. In the second stage, another segment $\boldsymbol{\tau}^{rf}$  is planned to connect the first path to the final goal $\mathbf{g}_f$ , thereby ensuring consistency with the long-term intention.
}
The pseudo-code of the intention-aware path replanning algorithm is given in Alg. \ref{alg1}.

\begin{algorithm}
\caption{Intention-aware Path Replanning Algorithm.}
\label{alg1}
\begin{algorithmic}
\State \textbf{Given:}
\begin{itemize}
    \item Control cycles: $T$, $\delta$
    \item Contextual information: $\mathbf{M}$, $\mathbf{g}_f$
    \item User-selected parameters: $W$, $H^r$, $H^h$
\end{itemize}
\State 

\State \textbf{Initialize:}
\State $\mathbf{p} \leftarrow \text{getRobotCurrentPosition}()$
\State $\boldsymbol{\tau}^r \leftarrow \text{pathPlanning}(\mathbf{p}, \mathbf{g}_f)$
\State $\boldsymbol{\gamma}^h \leftarrow \emptyset $  \Comment{reset the history buffer to empty}
\State $a \leftarrow 0$ \Comment{discrete action with 0: do nothing, 1: replan}
\State

\Every{$\delta$ seconds}
\State $\mathbf{p} \leftarrow \text{getRobotCurrentPosition}()$
\State $\mathbf{u}^h \leftarrow \text{getHumanCmd}()$
\State $\boldsymbol{\gamma}^h \leftarrow \text{updateBuffer}(\mathbf{p}, \mathbf{u}^h) $
    \Every{$T$ seconds}
        \If{humanInvolved($\boldsymbol{\gamma}^h$)}
            \State $\tilde{\mathbf{M}} \leftarrow \text{getLocalMap}(\mathbf{p}, \mathbf{M})$
            \State $\tilde{\mathbf{g}}_f, \tilde{\boldsymbol{\gamma}}^h, \tilde{\boldsymbol{\tau}}^r \leftarrow \text{getTransformed}(\mathbf{p}, \mathbf{g}_f, {\boldsymbol{\gamma}}^h, {\boldsymbol{\tau}}^r)$
            \State $\begin{aligned}[t]
            a, h, r \leftarrow {} & \text{predictIntentionDomain}(\tilde{\mathbf{M}}, \tilde{\mathbf{g}}_f,\\
            & \tilde{\boldsymbol{\gamma}}^h, \tilde{\boldsymbol{\tau}}^r, a)
            \end{aligned}$
        \EndIf

        \If{$a = 1$}
            \State $\mathcal{D}_{I} \leftarrow \text{getIntentionDomain}(\boldsymbol{\gamma}^h, h, r)$
            \State $ \hat{\mathbf{g}}_s \leftarrow \text{getPredictedSubgoal}(\mathcal{D}_{I}, \mathbf{M}, \mathbf{g}_f)$
            \State $\boldsymbol{\tau}^{rs} \leftarrow \text{pathPlanning}(\mathbf{p}, \hat{\mathbf{g}}_s)$ \Comment{constrained to the intention domain}
            \State $\boldsymbol{\tau}^{rf} \leftarrow \text{pathPlanning}(\hat{\mathbf{g}}_s, \mathbf{g}_f)$
            \State $\boldsymbol{\tau}^r \leftarrow \boldsymbol{\tau}^{rs} + \boldsymbol{\tau}^{rf}$
        \EndIf
    \EndEvery
\EndEvery
\end{algorithmic}
\end{algorithm}

The inputs of the algorithm consist of three parts: 

\begin{itemize}
   
\item [1)] 
\textbf{Control cycles}: the low-level control cycle $\delta$ determines the update rate of the robot's position and user commands, while the high-level decision cycle $T$ determines the frequency of path replanning.
 
\item [2)]
\textbf{Contextual information}: the occupancy grid map $\mathbf{M}$ and the final goal $\mathbf{g}_f$.

\item [3)]
\textbf{User-selected parameters}: the egocentric map width $W$, the observation length of robot intention $H^r$, and the buffer length of historical path points $H^h$

\end{itemize}

During initialization, a path from the robot's current position to the final goal $\mathbf{g}_f$ is planned with resolution $\Delta d$ as the robot's initial desired path, and the historical path buffer is empty. In each control cycle $\delta$, the buffer is updated according to the rules described in the previous subsection.

\textcolor{modify}{In each decision cycle $T$, the system first checks the buffer length to determine whether the user has intervened in the robot's motion over a sufficiently long distance. If this condition is satisfied, the path update process is triggered. In some cases, the user may only intend to perform precise} \textcolor{modify}{motion fine-tuning rather than modifying the robot's entire desired path. To account for this, the algorithm first determines whether replanning is required. When necessary, a trained reinforcement learning agent predicts the geometric parameters of the intention domain. According to (\ref{IntentD1}) and (\ref{IntentD2}), the intention domain is constructed in the local human frame and then transformed into the global map frame. \textcolor{second}{Then, among the collision-free points on the base of the cone-shaped intention domain, the one closest to the final goal $\mathbf{g}_f$ is selected as the predicted sub-goal $\hat{\mathbf{g}}_s$.} The first path segment $\boldsymbol{\tau}^{rs}$ is planned from the robot's current position to this sub-goal.}

The formulation of the RL problem and the process of policy training will be detailed in the next section.

\subsection{Reinforcement Learning Setup}

\textcolor{modify}{We formulate intention-aware path replanning as a sequential decision-making problem, where the robot acts as the agent and the human is treated as part of the environment. The problem is formalized as a Markov Decision Process (MDP) \cite{putermanMarkovDecisionProcesses2009}, defined by a 4-tuple $(\mathcal{S}, \mathcal{A}, \mathcal{R}, f)$, where $\mathcal{S}$ is the state space, $\mathcal{A}$ is the action space, $\mathcal{R} : \mathcal{S} \times \mathcal{A} \rightarrow \mathbb{R}$ is the reward function, and $f : \mathcal{S} \times \mathcal{A} \rightarrow \mathcal{S}$ denotes the transition dynamics.} An episode $(s_0, a_0, r_0, \dots, s_n, a_n, r_n)$ is a finite sequence of elements $s \in \mathcal{S}$, $a \in \mathcal{A}$, and $r \in \mathcal{R}$, beginning with a start state $s_0$ and ending when the environment terminates. We define a parameterized policy $\pi_{\psi} : \mathcal{S} \rightarrow \mathcal{A}$ with parameters $\psi$. The objective is to maximize $\mathbb{E}\:[\sum_{t=1}^{N}r(s_t, \pi_{\psi}(s_t))]$, which is the expected cumulative reward obtained in an episode under $\pi_{\psi}$.

\textcolor{modify}{The state is designed to capture contextual information regarding environmental structure, task priors, robot intention, and human intervention cues. It consists of five components: (i) a local cost map $\tilde{\mathbf{M}}$, cropped with width $W$ around the robot's current position; (ii) the final goal $\tilde{\mathbf{g}}_f$, represented in the local reference frame of the human; (iii) the robot's desired path segment $\tilde{\boldsymbol{\tau}}^r$ of length $H^r$; (iv) a historical path buffer $\tilde{\boldsymbol{\gamma}}^h$ containing $H^h$ past points; and (v) the most recent action type $a$. The trajectory segment $\tilde{\boldsymbol{\tau}}^r$ is extracted from the robot's planned path $\boldsymbol{\tau}^r$ by locating the point that is closest to the robot's current position and sampling $H^r$ subsequent points from it.}

The action space is a parameterized action space. Following notation in \cite{massonReinforcementLearningParameterized2016}, a complete action is represented as a tuple $(a, \mathbf{x}_a)$, where $a \in \mathcal{A}_d$ is the discrete action and $\mathbf{x}_a \in \mathcal{X}_a$ is the parameter executed with action $a$. The whole action space $\mathcal{A}$ is then the union of each discrete action with all possible parameters for that action:
\begin{equation}\label{p_action}
\begin{array}{l}
\mathcal{A} = \bigcup_{a \in \mathcal{A}_a} \{(a,\mathbf{x}_a)\:|\:\mathbf{x}_a\in\mathcal{X}_a\}.
\end{array}
\end{equation}
\textcolor{modify}{In our problem, the discrete action decides whether to trigger path replanning. If triggered, the associated parameters correspond to the geometric attributes of the intention domain, specifically the height $h$ and radius $r$; otherwise, no parameters are applied.}

\textcolor{modify}{One algorithm for solving this problem is the Hybrid Proximal Policy Optimization (H-PPO) \cite{fanHybridActorCriticReinforcement2019}. It adopts a hybrid actor-critic architecture and employs PPO \cite{schulmanProximalPolicyOptimization2017} for policy optimization. Detailed descriptions of the network architecture are provided in Appendix~\ref{app:network}.}

\textcolor{modify}{The training procedure and reward design are described in the following section.}

\subsection{Training with Human Trajectory Generation}

As shown in Alg. \ref{alg1}, the intention-aware path replanning is triggered only when the human intervenes to adjust the robot's motion. However, training such a policy directly with human participation is impractical, as the limited sampling efficiency of reinforcement learning would lead to prohibitively long training times. To overcome this challenge, we propose a Voronoi-based trajectory generation algorithm that synthesizes human-like trajectories online. This approach eliminates the need for human involvement during training and substantially improves training efficiency.

\textcolor{modify}{As discussed earlier, even when the final goal positions are specified, the existence of sub-goals yields multiple possible trajectories.} Therefore, we classify this trajectory uncertainty into global uncertainty and local uncertainty. 

Global uncertainty arises from the trajectory's topological relation to obstacles. For example, a human navigating around an obstacle may choose either the left or the right side. This form of uncertainty is naturally captured by homotopically distinct trajectories \cite{demeesterUseradaptedPlanRecognition2008}, which we model using Voronoi graphs. Local uncertainty stems from unknown sub-goal positions and the variability of human control. To capture it, we extend the concept of flight corridors \cite{renBubblePlannerPlanning2022}. Specifically, we build a corridor sequence composed of circles aligned with Voronoi edges, then sample via-points within each circle. Finally, a trajectory planner generates human-like trajectories by jointly considering obstacles, via-points, and the robot's kinematic constraints. The pseudo-code of the human trajectory generating algorithm is given in Alg. \ref{alg2}.

\begin{figure}
    \centering
    \includegraphics[width=8.3cm]{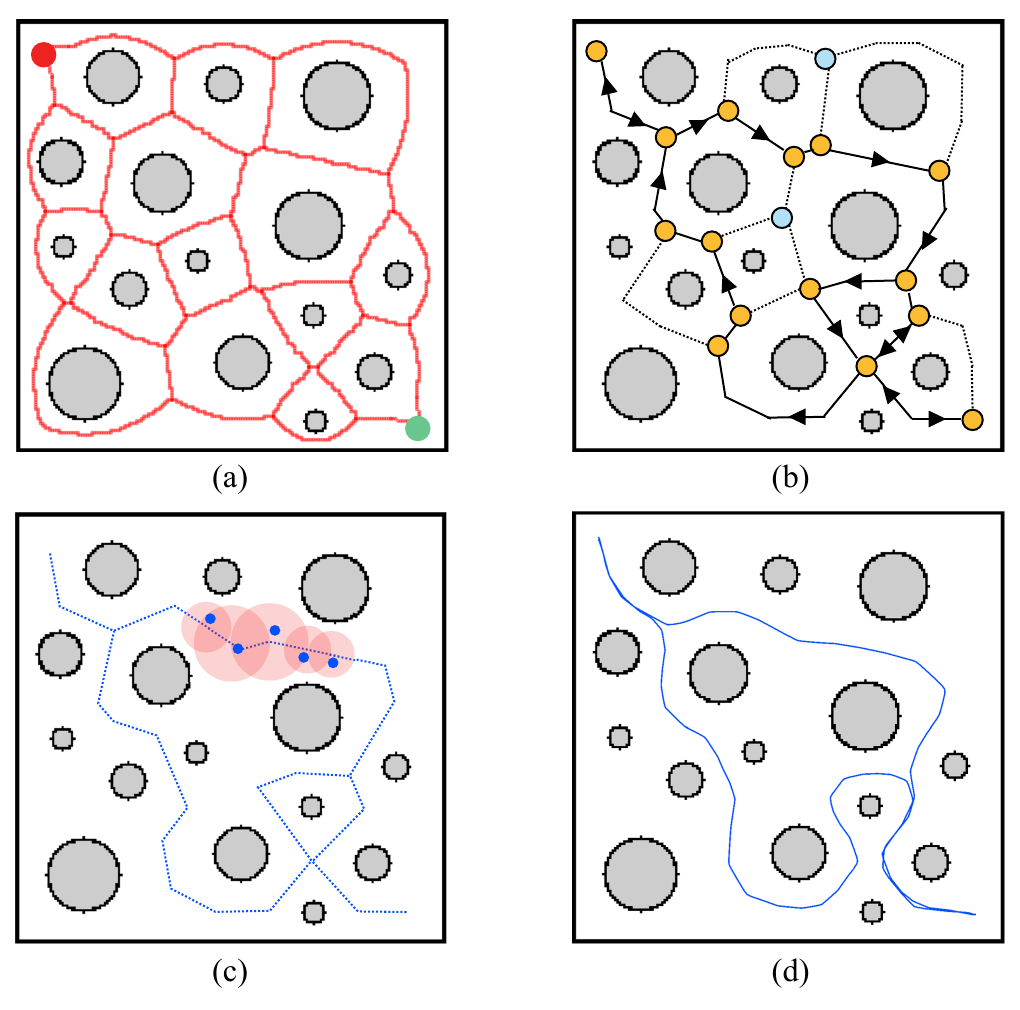}
    \caption{Voronoi-based human trajectory generation algorithm: (a) Extract the Discretized Voronoi Diagram (DVD) from the occupancy grid map. The start and goal are denoted by the green and red points. (b) Convert DVD into a Voronoi Graph. The sampled points are indicated by yellow nodes. (c) Generate initial trajectory along the Voronoi Graph edges, construct circular corridors and sample via-points within each corridor. (d) Trajectory planning under both corridor and via-point constraints}
    \label{traj_generator}
\end{figure}


\begin{algorithm}
\caption{Voronoi-based Human Trajectory Generation.}
\label{alg2}
\begin{algorithmic}
\State \textbf{Given:}
\begin{itemize}
    \item Global map: $\mathbf{M}$
    \item Start and end points: $\mathbf{p}_{start}$, $\mathbf{p}_{end}$
    \item Control cycle: $\delta$
\end{itemize}

\State 
\State \textbf{Initialize:}
\State $\mathbf{M}_v \leftarrow \text{getDiscretizedVoronoiDiagram}(\mathbf{M})$
\State $G \leftarrow \text{buildVoronoiDiagram}(\mathbf{M}_v, \mathbf{p}_{start}, \mathbf{p}_{end})$
\State $V_s \leftarrow \text{getStartNode}(G)$
\State $V_e \leftarrow \text{getEndNode}(G)$
\State 

\State $S \leftarrow \emptyset$ \Comment{the set of nodes on a simple path}

\State $S \text{.pushBack}(V_e)$
\State $V_{next} \leftarrow \text{sampleNeighbors}(G, V_e)$
\While{$V_{next} \neq V_s$}
    \State $S \text{.pushBack}(V_{next})$
        \If{hasNoValidNeighbor($G, V_{next}$)}
            \State $V_{next} \leftarrow \text{backTracking}(G, S)$
        \EndIf
        \State $V_{next} \leftarrow \text{sampleNeighbors}(G, V_{next})$
\EndWhile
\State $S \text{.pushBack}(V_s)$
\State $V_{next} \leftarrow \text{sampleNeighbors}(G, V_s)$
\While{$V_{next} \neq V_e$}
    \State $S \text{.pushBack}(V_{next})$
        \If{hasNoValidNeighbor($G, V_{next}$)}
            \State $V_{next} \leftarrow \text{backTracking}(G, S)$
        \EndIf
        \State $V_{next} \leftarrow \text{sampleNeighbors}(G, V_{next})$
\EndWhile
\State $S \text{.pushBack}(V_e)$

\State $\Gamma \leftarrow \text{getInitialPath}(G, S)$
\State $\mathcal{B} \leftarrow \text{getCorridor}(\Gamma)$
\State $\mathcal{P} \leftarrow \text{sampleViaPoints}(\mathcal{B})$
\State $\boldsymbol{\tau}^h \leftarrow \text{trajectoryPlanning}(\mathbf{p}_{start}, \mathbf{p}_{end}, \mathcal{B}, \mathcal{P}, \delta)$

\end{algorithmic}
\end{algorithm}

As shown in Fig. \ref{traj_generator} (a), the algorithm begins by converting the global occupancy grid into a Discretized Voronoi Diagram (DVD) \cite{lauEfficientGridbasedSpatial2013b}, defined as the set of points equidistant from at least two obstacles. Following \cite{kudererOnlineGenerationHomotopically2014}, the DVD is then converted into a Voronoi Graph consisting of nodes and edges, constructed with respect to the given start and goal positions. Edges that are too narrow for the robot to safely pass through are discarded. Since paths belonging to different homotopy classes may share common edges, we restrict sampling to simple paths from the Voronoi Graph. Furthermore, to model trajectory segments where the operator may temporarily move away from the global goal during sub-goal exploration, we sample a loop path by connecting a simple path sampled from the goal to the start with another path from the start to the goal, as illustrated in Fig. \ref{traj_generator} (b). Paths are sampled using a probabilistic depth-first search, selecting the next unvisited node uniformly and applying backtracking when necessary. For each sampled path, an initial trajectory is generated along the corresponding Voronoi edges. Circular corridors are then constructed around path points, where the radius equals the minimum distance from obstacles minus the robot radius. The center of the next circle is defined as the first path point outside the current circle. Within each corridor, a two-dimensional Gaussian distribution, parameterized by the circle center and radius, is used to sample via-points, as shown in Fig. \ref{traj_generator} (c). Finally, the Timed Elastic Band (TEB) trajectory planning algorithm \cite{7324179} is used to generate the human-like trajectory. 

At the start of each training episode, a human trajectory is generated and the robot moves along this trajectory. Once the length of the historical path buffer reaches $H^h$, the path replanning module (Alg. \ref{alg1}) is periodically executed at intervals of $T$. At each step, the agent determines whether to trigger replanning and, if so, how to construct the intention domain. If no feasible path is found within the intention domain, the robot halts and remains in the same state until a feasible action is obtained. The maximum episode length is limited based on the size of the map.

The reward function consists of three components:
\begin{equation}
\label{reward}
\begin{array}{l}
R = w_1 r_{task} + w_2 r_{terminal} + r_{reg}.
\end{array}
\end{equation}
In (\ref{reward}), $r_{task}$ encourages the replanned trajectory to align with the human-intended path. This reward is inversely proportional to the Euclidean distance error between the next $H^r$ points of the human-desired and robot-desired paths. The exponential term normalizes the error to the range $[0, 1]$, while a geometric weighting assigns higher importance to points closer to the robot's current position $\mathbf{p}$:
\begin{equation}\label{task_reward}
\begin{array}{l}
r_{task} = \frac{1 - \lambda}{1 - \lambda^{H^r}} \sum_{j=1}^{H^r} \lambda^{j-1} e^j \\
e^j = \exp (-\eta || \mathbf{p}^r(k^r + j) - \mathbf{p}^h(k^h + j) ||_2)\\
k^i = \text{argmin}_{1 \le k \le |\boldsymbol{\tau}^i|} ||\mathbf{p}^i(k) - \mathbf{p}||, \quad i=\{r, h\}.
\end{array}
\end{equation}
\textcolor{modify}{The task reward encourages the robot desired path to stay close to the future human-intended path. A smaller distance between the two paths leads to a reward closer to 1.}

\textcolor{modify}{The term $r_{terminal}$ is a sparse reward that penalizes invalid intention-domain parameters and rewards successful task completion:}
\begin{equation}\label{terminal_reward}
\begin{array}{l}
r_{terminal} = \left\{
\begin{array}{ll}
+1 ,&  \text{if} \:\:\: \text{final goal reached}    \\
-1 ,& \text{if} \:\:\: \text{intention domain invalid}    \\
0 ,& \text{otherwise.}   \
\end{array}
\right.
\end{array}
\end{equation}
\textcolor{modify}{Invalid intention-domain parameters can make replanning infeasible. For example, when the entire base of the intention domain lies inside obstacles, the replanning algorithm cannot find a valid sub-goal on the base and thus fails to generate a feasible path.} The weights $w_1$ and $w_2$ balance the contributions of the task reward and the terminal reward.

Finally, $r_{reg}$ is a regularization reward that constrains the geometric shape of the intention domain:
\begin{equation}\label{reg_reward}
\begin{array}{l}
r_{reg} = \left\{
\begin{array}{ll}
\beta \frac{\ln(1-\phi_{norm})} {1-\phi_{norm}} ,&  \text{if} \:\:\: r_{reg} \ge  -w_2  \\
-w_2 ,& \text{otherwise.}   
\end{array}
\right.
\end{array}
\end{equation}
When the cone base overlaps with obstacles, enlarging the base radius no longer affects the sub-goals; thus, wider cone angles are penalized. The penalty increases with the normalized half-opening angle $\phi_{norm}$ and is scaled by parameter $\beta$, but its magnitude is bounded by $-w_2$ to maintain training stability. \textcolor{modify}{Since this reward is negative, it not only influences the average opening angle of the intention domain but also penalizes each replanning decision, making replanning triggers less sensitive.}

\textcolor{modify}{\textbf{Remark 1:} To address the gap between synthetic training data and real human behavior, our algorithm maximizes the diversity of the generated trajectories. By selecting appropriate starting positions, the generated trajectories cover every directed edge on the Voronoi graph, effectively capturing global trajectory uncertainty across all possible human intentions. At the local level, when sampling via-points, we assume humans naturally prefer safer paths farther from obstacles. This assumption holds in most cases for the following reasons. The robot maintains a safe distance when following its planned trajectory autonomously. When the human intervenes, the safety constraints provided by the local shared controller ensure continued obstacle clearance. Furthermore, our framework is planner-agnostic; it does not strictly rely on the TEB trajectory planner and can be integrated with any planning method that satisfies obstacle avoidance, via-point, and kinematic constraints.}

\section{Locally Shared Controller}

In contrast to the training process, in real interactions the agent does not need to trigger path replanning when the robot's intended path is already consistent with the human's intention. 
To accommodate such interventions while maintaining safety, we implement a locally shared controller, drawing inspiration from shared DWA \cite{6840152}.

The Dynamic Window Approach (DWA) \cite{580977} remains one of the most widely used methods for local obstacle avoidance in mobile robot navigation. By sampling candidate velocity commands within the dynamic velocity window and minimizing a multi-objective cost function, DWA enables safe, smooth, and goal-directed motion in unstructured environments. To adapt DWA for shared control, prior works \cite{6840152} modify the cost function to incorporate user input. Following this idea, we design a switched local controller. When the operator provides no input, the controller defaults to a standard DWA planner. When the operator issues a command, the controller first evaluates its feasibility by computing clearance, defined as the distance from the terminal point of the commanded trajectory to the closest obstacle. If the clearance exceeds a predefined threshold, the optimal action is chosen by minimizing a modified cost function:
\begin{equation}\label{dwa_cost}
\begin{array}{l}
C = 1 - Clearance + Clearance\cdot Cost_{cmd} \\
Cost_{cmd} = 0.5 \cdot Heading + 0.5 \cdot Velocity,
\end{array}
\end{equation}
where the Heading term measures how well the curvature fits the human target curvature, and the Velocity term assesses how close the linear speed $v$ is to the user's desired linear speed $v^h$. All cost terms are normalized to $[0,1]$.
  
\section{Simulations}
\label{sim_section}

\textcolor{modify}{To simulate cluttered environments, we randomly generated two $10 \text{m} \times 10 \text{m}$ maps with different types of obstacles. As shown in Fig. \ref{sim_setting}, one map contained non-overlapping circular obstacles, while the other contained rectangular obstacles. The start and goal positions were located at diagonally opposite corners of the map. Human desired trajectories were produced using Alg. \ref{alg2} with a sampling interval of  $0.1\, \text{s}$ and were employed for both model training and evaluation. For the intention-aware path replanning algorithm, the low-level control cycle was set to $\delta =0.1\, \text{s}$ and the high-level decision cycle was set to $T = 0.5\,\text{s}$. The egocentric map had a width of $W = 5\,\text{m}$; the observation horizon of robot intention and the length of the historical buffer were $H^r = 100$ and $H^h = 20$, respectively. The path planner employed the wavefront Dijkstra algorithm. The reward function employs $\eta = 1.0$ and $\lambda = 0.98$, with corresponding weights $w_1 = 2$ for the task reward and $w_2 = 2$ for the terminal reward. Detailed training settings and network architecture are provided in the appendix (see Appendix \ref{app:network}).}

\subsection{Evaluation with Generated Human Trajectories}

\textcolor{modify}{To validate our method for replanning the robot desired path to align with human short-term intentions, and to analyze the effects of the regularization and terminal reward terms, we evaluated its performance with generated human trajectories in the map with circular obstacles. We extracted all simple paths connecting the start and goal nodes from the Voronoi graph. Following on Alg. \ref{alg2}, we generated looped trajectories with the same forward and reverse simple paths. This procedure produced a total of 256 trajectories, comprehensively capturing global trajectory uncertainty. The generated trajectories served as ground truth for the human desired path. Since no human operator was involved, the robot followed the generated trajectories exactly, as if controlled by a virtual human.}


\textcolor{modify}{We adopted a commonly used path replanning method, periodic global replanning, as the baseline (PG-Baseline). This approach constantly recalculates the global path at a fixed frequency to adapt to unpredictable environments. In our setting, it recomputed the entire path from the current robot position to the final goal at every high-level decision interval, generating a new robot desired path. Both methods employed Dijkstra's algorithm to compute the path based on the global costmap.}

\textcolor{modify}{For each trajectory, the Mean Euclidean Distance (MED) between the robot-desired and human-desired paths was computed to evaluate prediction error $e_{MED}$. As the robot made decisions periodically, MED was calculated over the next $H^r$ points at each decision step and averaged across the entire run. To observe the learned behaviors, we also recorded the replanning frequency $f$, the average opening angle $2\phi$ of the predicted intention domains and failure rate $r_{fail}$. The replanning frequency was defined as the ratio of successful replanning events to the total decision steps, while the failure rate was defined as the ratio of failed events to the total decision steps. A Failure occurred when the predicted intention domain was invalid, such as when the predicted sub-goal cannot be identified or when no feasible path is found during replanning. Final results were reported as the mean values over all trajectories.}

\textcolor{modify}{Table \ref{tab:pure_sim} summarizes the quantitative results. Compared with the PG-Baseline, our method greatly reduced the replanning frequency while achieving lower prediction errors, indicating that the predicted intention domain effectively captured human short-term intentions. The lower replanning frequency further suggested that our approach could recognize local uncertainties in the human trajectory and maintain stable robot behavior despite operational noise or minor interventions.}

\textcolor{modify}{As $\beta$ in the regularization reward term increased, both the replanning frequency and the opening angle of the intention domain decreased, while the failure rate remained stable at approximately 1.00\%. This demonstrated that the regularization reward effectively shapes the replanning strategy. A smaller opening angle resulted in the first segment of the replanned path being more aligned with the human's current direction, while less frequent replanning implied that the robot's intention was less sensitive to the uncertainty of human operation. After introducing the terminal reward term, the failure rate dropped significantly, demonstrating its effectiveness in reducing failure events. Further increases resulted in only a marginal decrease in the failure rate, with negligible changes in other evaluation metrics. Therefore, we fixed the terminal reward weight at $w_2 = 2.0$ and selected $\beta$ based on the average prediction error. According to Table \ref{tab:pure_sim}, $\beta = 0.5$ achieved the lowest average prediction error and was thus used in subsequent evaluations.}

\begin{figure}[t]
    \centering
    \includegraphics[width=8.5cm]{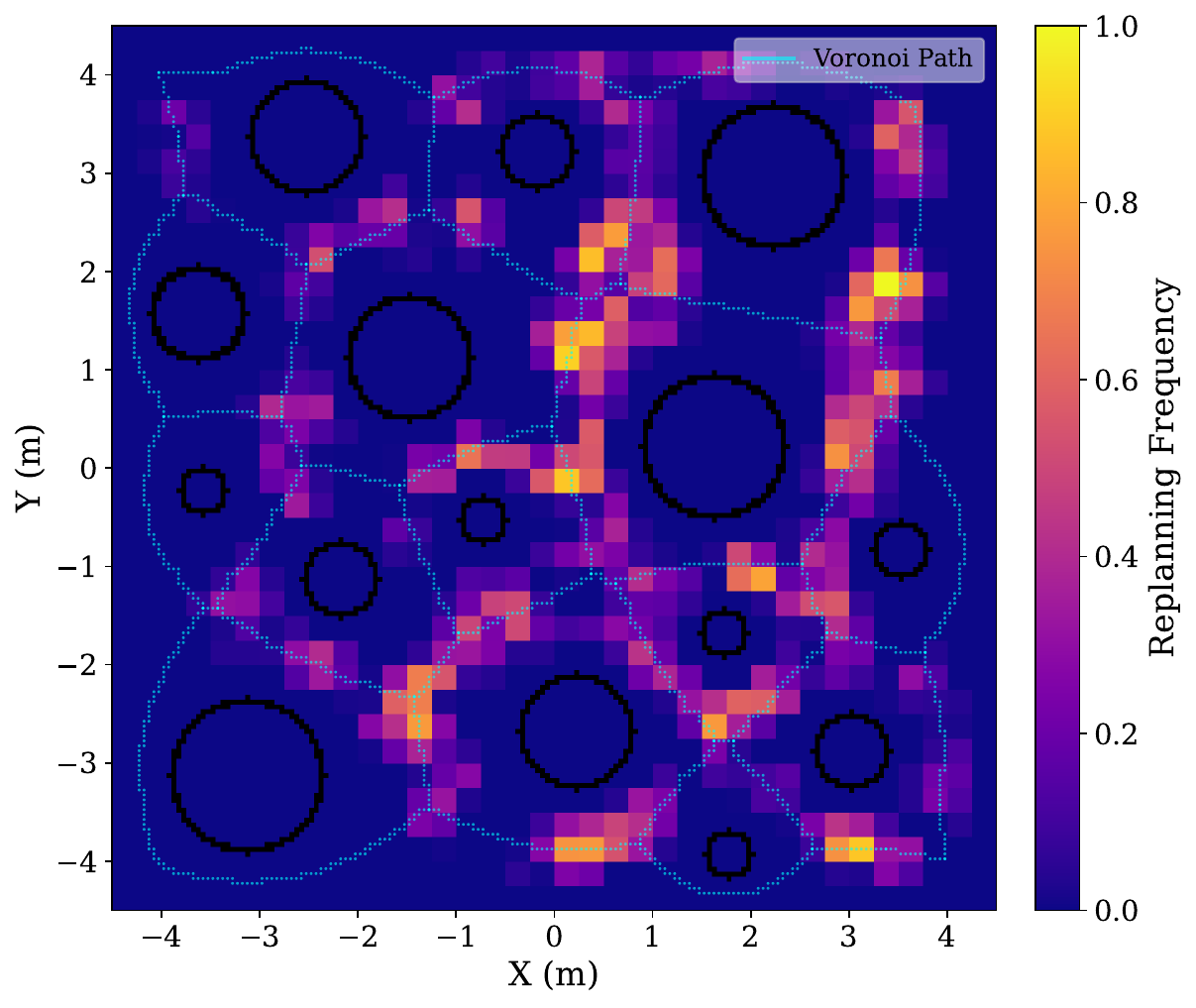}
    \caption{Heat map visualization of the spatial distribution of replanning frequency, evaluated over 256 human-desired trajectories. The discretized Voronoi diagram is overlaid in green lines.}
    \label{heatmap}
\end{figure}

\textcolor{modify}{To better understand the learned path replanning strategy, we examined the spatial distribution of replanning frequency. Fig. \ref{heatmap} illustrated the spatial distribution of replanning frequency for the model with the lowest prediction error, evaluated over 256 human-desired trajectories. Replanning was rarely triggered when the robot was near Voronoi nodes but became more frequent along edges. This occurred because, at nodes, the agent faced high global and local uncertainty regarding the human's intended direction, whereas along edges, surrounding obstacles reduced the global uncertainty, making the human intentions more predictable.}

\begin{table}[t]
  \centering
  \caption{Evaluation with Generated Human Trajectories. Results are averaged over three runs with different random seeds.}
  \label{tab:pure_sim}
  \setlength{\tabcolsep}{0pt} 
  \begin{tabular*}{\columnwidth}{@{\extracolsep{\fill}}l|cccc}
    \toprule
    Model & $e_{MED}$/m ($\downarrow$) & $f$ & $2\phi$ / $^\circ$ & $r_{fail}$/\% ($\downarrow$) \\
    \midrule
    PG-Baseline       & $1.32 \pm 0.06$ & $1.00 \pm 0.00$ & - & - \\
    $\beta = 0.0$  & $0.96 \pm 0.11$ & $0.29 \pm 0.04$ & $92.41 \pm 7.76$ & $1.16 \pm 0.70$ \\
    $\beta = 0.1$  & $0.97 \pm 0.12$ & $0.27 \pm 0.03$ & $69.18 \pm 6.22$ & $1.03 \pm 0.68$ \\
    $\beta = 0.5$  & $\mathbf{0.88 \pm 0.11}$ & $0.20 \pm 0.03$ & $56.22 \pm 6.64$ & $1.09 \pm 0.68$ \\
    $\beta = 2.0$  & $0.93 \pm 0.12$ & $0.17 \pm 0.03$ & $33.46 \pm 5.98$ & $1.42 \pm 0.79$ \\
    \midrule 
    $w_2 = 0.0$    & $1.28 \pm 0.17$ & $0.41 \pm 0.06$ & $90.74 \pm 10.51$ & $5.97 \pm 1.89$ \\
    $w_2 = 2.0$    & $\mathbf{0.88 \pm 0.11}$ & $0.20 \pm 0.03$ & $56.22 \pm 6.64$ & $1.09 \pm 0.68$ \\
    $w_2 = 5.0$    & $0.93 \pm 0.13$ & $0.16 \pm 0.02$ & $53.56 \pm 6.94$ & $0.54 \pm 0.49$ \\
    $w_2 = 10.0$   & $0.89 \pm 0.11$ & $0.19 \pm 0.03$ & $47.90 \pm 5.01$ & $0.40 \pm 0.43$ \\
    \bottomrule
  \end{tabular*}
\end{table}

\subsection{Evaluation with Human in the Loop}
\label{sec_sim}

\begin{figure}[t]
    \centering
    \includegraphics[width=8.5cm]{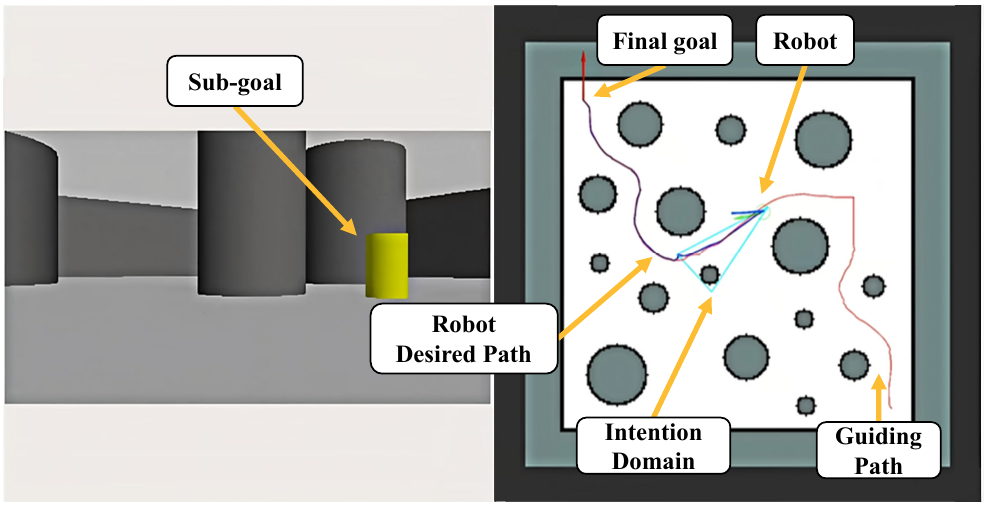}
    \caption{Monitor interface in simulation. Left: the robot's front camera view. Right: the occupancy grid map with the robot's desired trajectory, final goal and guiding path.}
    \label{monitor}
\end{figure}

\begin{figure}[t]
    \centering
    \includegraphics[width=8cm]{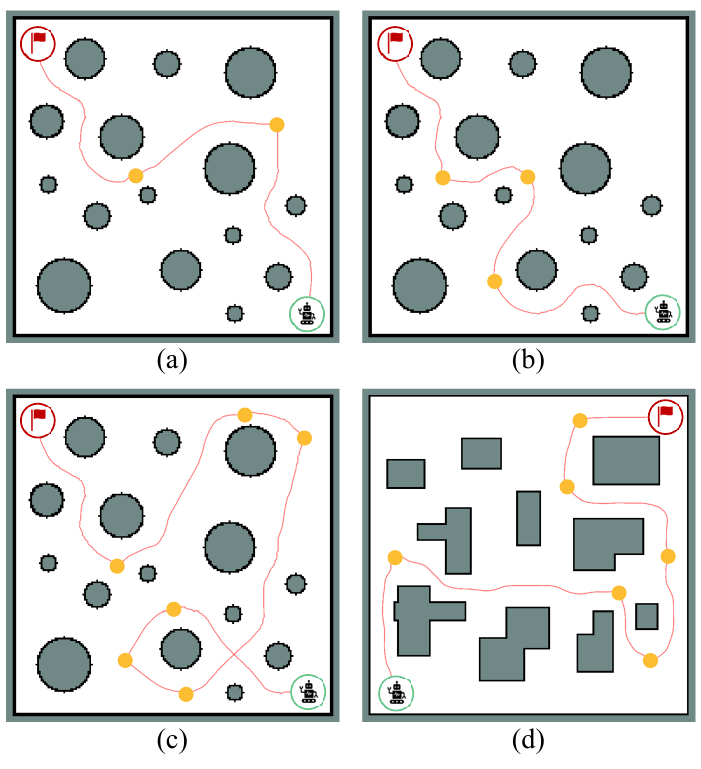}
    \caption{\textcolor{modify}{Visualization of four sub-goal distribution scenarios.}}
    \label{sim_setting}
\end{figure}

\textcolor{modify}{To rigorously analyze the behavior of the agent in a controlled and repeatable setting, we performed a human-in-the-loop simulation in Gazebo. The simulation environment was constructed based on the occupancy grid map shown in Fig. \ref{sim_setting}, using the same start and goal positions as in the prior experiment. The robot platform was TurtleBot3. Localization was achieved through the Nav2 framework \cite{macenski2020marathon2}. The sub-goals were represented as yellow cylinders. To inform the operator of the sub-goal distributions, a guiding path connecting all sub-goals sequentially from start to goal was overlaid on the occupancy grid map. This design explicitly conveyed the intended route to the operator, ensuring consistent motion intent during evaluation. The human operator controlled the robot via a joystick. As shown in Fig. \ref{monitor}, the monitoring interface displayed both the front camera view (left) and the occupancy grid map (right). The camera view enabled the operator to identify sub-goal positions, while the map provided visualization of the robot's current planned trajectory and the guiding path.}

\textcolor{modify}{To represent diverse human intentions, three sub-goal distributions were defined in the map with circular obstacles (Fig. \ref{sim_setting} (a)-(c)), and one distribution was defined in the map with rectangular obstacles (Fig. \ref{sim_setting} (d)). From case (a) to (c), the number of sub-goals increased. From case (c) to (d), the number of sub-goals remained unchanged, while the operational difficulty increased due to the presence of narrow passages in the map with rectangular obstacles.}

\textcolor{modify}{To further verify the effectiveness of our approach (SC) in reducing operator effort without compromising task efficiency, we adopted the shared DWA algorithm as a baseline, which required direct human control (HL). Under this method, the robot assisted the operator in obstacle avoidance but did not perform autonomous motion planning.} For our method, the policy with the lowest prediction error, as reported in Table \ref{tab:pure_sim}, was selected for this experiment. Algorithm performance was evaluated using the following objective metrics:

{\color{modify}
\begin{itemize}

\item [1)] 
\textbf{Completion time} $T_{c}$: total time required to complete the task.

\item [2)] 
\textbf{Interaction time} $T_{h}$: cumulative time during which the human provided control input.

\item [3)]
\textbf{Interaction percent} $\rho$: ratio of the interaction time to the completion time.

\item [4)]
\textbf{Trajectory clearance} $D_{clear}$: average distance from each trajectory point to the nearest obstacle.

\end{itemize}
}

\textcolor{modify}{\textcolor{second}{For each case under each control condition, three independent runs were conducted, and the mean and standard deviation of each metric are reported in Table~\ref{tab:sim}. Our method substantially reduced the interaction percentage in all four cases.} Although the interaction percentage increased slightly as the number of sub-goals increased, our method still reduced nearly half of the human effort in the case (c). Additionally, trajectory safety improved consistently under all conditions. These results demonstrated that our method effectively reduced the operator's workload and enhanced safety. A slight increase in task completion time was observed in all three cases within the map containing circular obstacles. This was attributed to two factors. First, the circular map required less precise manual control from the operator compared to the rectangular map. Second, before transferring control authority back to the robot, the operator required additional time to perceive and assess the updated robot's desired path. Therefore, as task conditions become more demanding, such as with higher control complexity or stricter safety requirements, the advantages of our method are expected to become even more pronounced.}

To better understand the path replanning strategy, we visualized the full cooperation process between the human and robot from a representative experiment under one sub-goal distribution case (Fig. \ref{sim_setting} (c)).

\begin{figure}
    \centering
    \includegraphics[width=8.5cm]{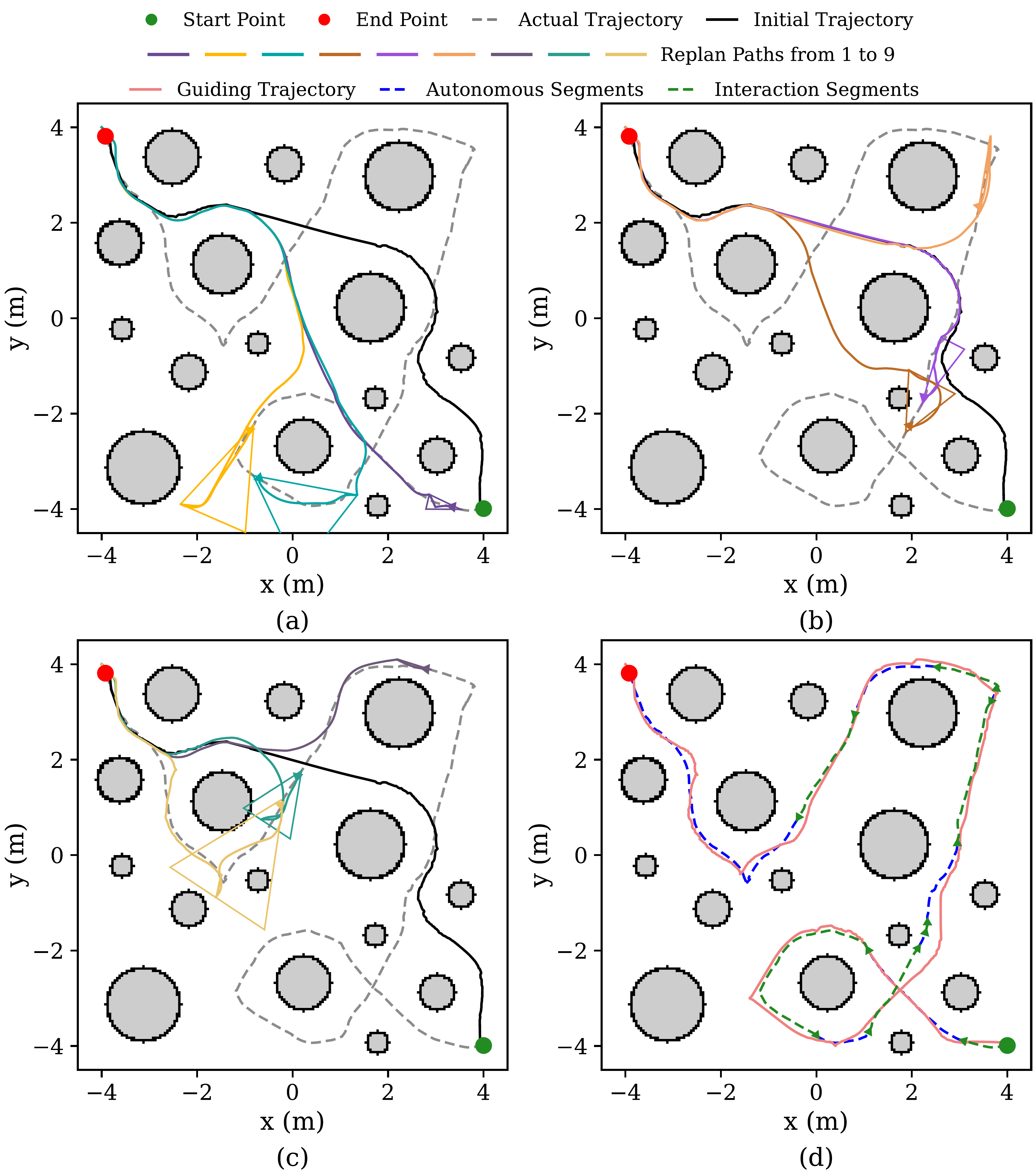}
    \caption{\textcolor{modify}{(a)-(c): Replanned paths and corresponding intention domains from one representative experiment (nine in total). (d): Actual versus guiding trajectory, with human-intervened segments shown in green.}}
    \label{replanning_process}
\end{figure}

\textcolor{modify}{As shown in Fig. \ref{replanning_process}, the gray dashed line represented the actual motion trajectory and the black solid line denoted the robot's initially planned path. \textcolor{third}{Fig. \ref{replanning_process} (a)-(c) illustrated the nine replanning events during the experiment along with their corresponding intention domains.} Among them, the second, sixth, and eighth replanned paths did not change the homotopy relationship of the robot's desired path with respect to obstacles. This was because these triggers occurred in relatively open areas, where the local uncertainty of the human's intended trajectory was greater, leading the robot to focus more on the human's current direction during replanning. Moreover, the third, fourth, and fifth replanning results showed that when multiple homotopically distinct trajectories were available, the agent tended to select the one with the shortest overall length. This behavior was consistent with the trajectory generation strategy in Alg. \ref{alg2}, where shorter paths had a higher probability of being sampled. Subfigure (d) compared the actual motion trajectory with the guiding path, marking the segments where human interventions occurred. Subfigure (d) also showed that human intervention occurred more frequently in open regions, while control authority was returned to the robot in narrow spaces.}

\begin{table}[t]
  \centering
  \caption{\textcolor{second}{Evaluation results with human in the loop under four different sub-goal distributions.}}
  \label{tab:sim}
  \setlength{\tabcolsep}{3pt}
  \renewcommand{\arraystretch}{1.05}

  \resizebox{\columnwidth}{!}{%
  \begin{tabular}{@{}ll|cccc@{}}
    \toprule
    Case & Condition
    & $T_c$ / s ($\downarrow$)
    & $T_h$ / s ($\downarrow$)
    & $\rho$ ($\downarrow$)
    & $D_{\mathrm{clear}}$ / m ($\uparrow$) \\
    \midrule

    (a) & HL
        & $\mathbf{31.05 \pm 1.31}$
        & $30.30 \pm 0.89$
        & $97.65 \pm 2.19\%$
        & $0.54 \pm 0.03$ \\
        & SC
        & $33.87 \pm 0.90$
        & $\mathbf{8.20 \pm 1.69}$
        & $\mathbf{24.22 \pm 5.27}\%$
        & $\mathbf{0.61 \pm 0.00}$ \\
    \midrule

    (b) & HL
        & $\mathbf{34.94 \pm 1.69}$
        & $34.38 \pm 1.61$
        & $98.39 \pm 0.13\%$
        & $0.54 \pm 0.02$ \\
        & SC
        & $38.83 \pm 0.58$
        & $\mathbf{15.37 \pm 0.36}$
        & $\mathbf{39.59 \pm 0.49}\%$
        & $0.54 \pm 0.01$ \\
    \midrule

    (c) & HL
        & $\mathbf{64.35 \pm 4.21}$
        & $63.65 \pm 4.22$
        & $98.91 \pm 0.07\%$
        & $0.55 \pm 0.03$ \\
        & SC
        & $67.97 \pm 2.11$
        & $\mathbf{33.51 \pm 3.49}$
        & $\mathbf{49.35 \pm 5.66}\%$
        & $\mathbf{0.58 \pm 0.02}$ \\
    \midrule

    (d) & HL
        & $67.75 \pm 3.95$
        & $67.56 \pm 3.97$
        & $99.11 \pm 0.60\%$
        & $0.46 \pm 0.00$ \\
        & SC
        & $\mathbf{64.88 \pm 4.39}$
        & $\mathbf{20.49 \pm 3.59}$
        & $\mathbf{31.49 \pm 4.06}\%$
        & $\mathbf{0.50 \pm 0.01}$ \\
    \bottomrule
  \end{tabular}%
  }
\end{table}

\subsection{Justification of the Intention Domain Shape}\label{sec_GeneraID}

\begin{figure*}
    \centering
    \includegraphics[width=16.5cm]{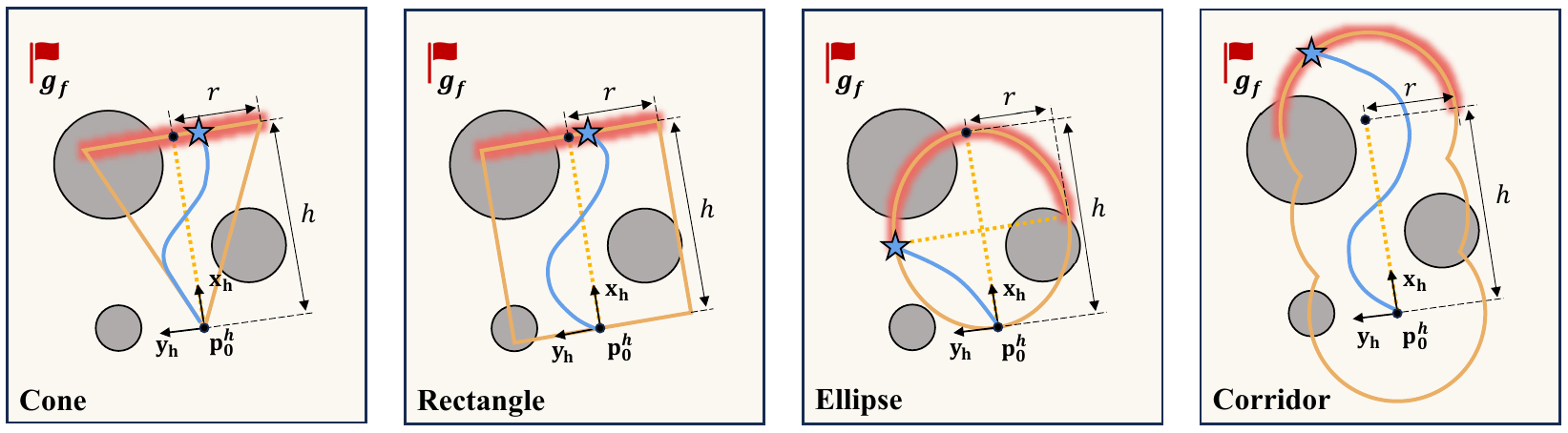}
    \caption{\textcolor{modify}{Visualization of the four intention domain shapes: Cone, Rectangle, Ellipse, and Corridor. All shapes are defined in the human's local reference frame with identical height $h$ and radius $r$. The forward boundaries, highlighted in red, indicate candidate regions for computing the predicted sub-goal (blue star). This sub-goal is selected as the collision-free point on the boundary that is closest to the final goal $g_f$ (red flag). The blue solid line illustrates the replanned path segment navigating to the predicted sub-goal.}}
    \label{intention_domains}
\end{figure*}

\begin{table*}[t]
  \centering
  \caption{\textcolor{second}{Comparative performance of different intention domain shapes in the human-in-the-loop simulation under Case (c).}}
  \label{tab:intention domain shape}
  \setlength{\tabcolsep}{4pt}
  \renewcommand{\arraystretch}{1.05}

  \resizebox{\linewidth}{!}{%
  \begin{tabular}{@{}l|cccccccc@{}}
    \toprule
    Shape
    & $T_{c}$ / s ($\downarrow$)
    & $T_{h}$ / s ($\downarrow$)
    & $\rho$ ($\downarrow$)
    & $\Delta T_{c}$ ($\downarrow$)
    & $\Delta T_{h}$ ($\downarrow$)
    & $D_{\mathrm{clear}}$ / m ($\uparrow$)
    & $N_{\mathrm{replan}}$ ($\downarrow$)
    & $T_{\mathrm{infer}}$ / ms ($\downarrow$) \\
    \midrule

    Rect
    & $81.80 \pm 6.68$
    & $52.49 \pm 10.49$
    & $63.85 \pm 8.09\%$
    & $27.12 \pm 10.38\%$
    & $-17.53 \pm 16.48\%$
    & $0.57 \pm 0.02$
    & $12.67 \pm 1.53$
    & $\mathbf{17.54 \pm 4.74}$ \\

    Elli
    & $80.91 \pm 2.27$
    & $42.82 \pm 7.63$
    & $52.78 \pm 8.01\%$
    & $25.75 \pm 3.52\%$
    & $-32.72 \pm 11.99\%$
    & $0.57 \pm 0.00$
    & $10.67 \pm 0.58$
    & $20.36 \pm 5.55$ \\

    Corr
    & $70.33 \pm 3.68$
    & $37.66 \pm 3.40$
    & $53.76 \pm 7.12\%$
    & $9.29 \pm 5.71\%$
    & $-40.82 \pm 5.24\%$
    & $\mathbf{0.59 \pm 0.01}$
    & $13.00 \pm 5.29$
    & $22.07 \pm 8.72$ \\

    Cone
    & $\mathbf{67.97 \pm 2.11}$
    & $\mathbf{33.51 \pm 3.49}$
    & $\mathbf{49.35 \pm 5.66}\%$
    & $\mathbf{5.63 \pm 3.29}\%$
    & $\mathbf{-47.35 \pm 5.49}\%$
    & $0.58 \pm 0.02$
    & $\mathbf{8.67 \pm 0.58}$
    & $18.57 \pm 3.46$ \\

    \bottomrule
  \end{tabular}%
  }
\end{table*}

As mentioned in Section \ref{ID_sec}, the intention domain is defined as a non-empty, connected, and closed set in $\mathbb{R}^2$ and is modeled as a closed cone. To justify this choice, we further conducted a comparative experiment with other commonly used geometric representations. As shown in Fig. \ref{intention_domains}, three additional shapes were considered: rectangular, elliptical, and circular corridor domains. All configurations share the same number of geometric parameters, height $h$ and radius $r$. Similar to the cone-shaped intention domain, all shapes are defined in the local reference frame of the human. Let $\mathbf{q} = [x, y]^T \in \mathbb{R}^2$ denote an arbitrary point in this frame. The rectangular intention domain is defined as:
\begin{equation}\label{rect}
\begin{array}{l}
\tilde{\mathcal{D}}_I^{rec}\:(h, r) = \left\{ \mathbf{q} \in \mathbb{R}^2 \:\: \bigg| \:\: x \in [0, h], \:\: |y| \le r \right\},
\end{array}
\end{equation}
and the ellipse intention domain is defined as:
\begin{equation}\label{elli}
\begin{array}{l}
\tilde{\mathcal{D}}_I^{ell}\:(h,r) = \left\{ \mathbf{q} \in \mathbb{R}^2 \:\: \bigg| \:\: \frac{(x-h)^2}{h^2} + \frac{y^2}{r^2} \le 1\right\}.
\end{array}
\end{equation}
In motion planning, a safe corridor is typically defined as a series of connected, overlapping convex polyhedra that model the obstacle-free space between a robot and its goal \cite{liuPlanningDynamicallyFeasible2017a}. Since path planning under the intention domain constraint guarantees collision avoidance, additional safety constraints are unnecessary. Therefore, without increasing model complexity, the circular corridor intention domain is defined as:
\begin{equation}\label{corridor}
\tilde{\mathcal{D}}_I^{cor}(h,r)
=
\bigcup_{x_c \in [0,h]}
\left\{
\mathbf{q} \in \mathbb{R}^2
\mid
(x-x_c)^2 + y^2 \le r^2
\right\}.
\end{equation}
In practice, it is approximated using a finite number of circular regions. The predicted sub-goal is selected as the collision-free point on the forward boundary of the intention domain that is closest to the final goal. The forward boundary of each shape is highlighted in red in Fig. \ref{intention_domains}.

We trained the models using the same parameters and network architecture as the cone-shaped intention domain on the map with circular obstacles. The trained models were then evaluated in a human-in-the-loop setting under case (Fig. \ref{sim_setting} (c)). In addition to the evaluation metrics introduced in the previous subsection, we further considered the following metrics:

\begin{itemize}

\item [1)] 
\textbf{Normalized completion time increment} $\Delta T_{c}$:  proportion of the increase in completion time under shared control relative to the total completion time under direct human control.

\item [2)] 
\textbf{Normalized interaction time increment} $\Delta T_{h}$: proportion of the increase in interaction time under shared control relative to the total interaction time under direct human control.

\item [3)]
\textbf{Number of replanning events} $N_{replan}$:  total number of path replanning events.

\item [4)]
\textbf{Inference time} $T_{infer}$: average computation time for intention domain prediction and path replanning.

\end{itemize}

\textcolor{second}{For each intention domain shape, three independent runs were conducted, and the mean and standard deviation of each metric are reported in Table~\ref{tab:intention domain shape}.} The cone-shaped intention domain achieved the lowest task completion time and interaction time among all methods. It reduced interaction time by 47.35\% with only a 5.63\% increase in completion time. For trajectory safety, all four intention domain representations showed nearly identical performance. The cone-shaped intention domain also resulted in the fewest replanning events, indicating that it produced fewer wrong predictions that were inconsistent with the human's short-term intention, in addition to correct predictions. In terms of computational efficiency, all four representations achieved inference times of around $20 \:\text{ms}$, satisfying real-time requirements. Among them, the rectangular intention domain achieved the lowest inference time due to its simpler boundary computation, followed by the cone-shaped domain. Overall, these results validate the effectiveness of modeling the intention domain as a cone. 


Furthermore, when the intention domain is modeled as a cone, the cone height $h$ determines the prediction horizon of short-term human intention, while the base radius $r$ constrains the alignment between the predicted intention and the human's current direction. Given a fixed opening angle, the spatial region covered by the intention domain expands non-linearly with increasing prediction horizon. This geometric property naturally aligns with the characteristic that uncertainty in human motion intention increases over time. In contrast, under the same geometric parameters, alternative shapes such as rectangles enclose a larger spatial region, which reduces the likelihood of generating paths consistent with the true human intention during replanning.

\subsection{Comparative Analysis with Heuristic Baselines}
\label{sec_heu}

\begin{table*}[t]
  \centering
  \caption{\textcolor{second}{Comparative performance between the heuristic baseline and the RL method.}}
  \label{tab:heuristic}
  \setlength{\tabcolsep}{3.5pt}
  \renewcommand{\arraystretch}{1.05}

  \resizebox{\linewidth}{!}{%
  \begin{tabular}{@{}ll|cccccccc@{}}
    \toprule
    Case & Method
    & $T_c$ / s ($\downarrow$)
    & $T_h$ / s ($\downarrow$)
    & $\rho$ ($\downarrow$)
    & $\Delta T_c$ ($\downarrow$)
    & $\Delta T_h$ ($\downarrow$)
    & $D_{\mathrm{clear}}$ / m ($\uparrow$)
    & $N_{\mathrm{replan}}$ ($\downarrow$)
    & $T_{\mathrm{infer}}$ / ms ($\downarrow$) \\
    \midrule

    (c) & Heuristic
    & $72.68 \pm 2.36$
    & $34.31 \pm 1.27$
    & $\mathbf{47.07 \pm 2.17}\%$
    & $12.96 \pm 3.66\%$
    & $-46.09 \pm 2.00\%$
    & $0.57 \pm 0.01$
    & $21.67 \pm 1.15$
    & $22.21 \pm 2.88$ \\

        & RL
    & $\mathbf{67.97 \pm 2.11}$
    & $\mathbf{33.51 \pm 3.49}$
    & $49.35 \pm 5.66\%$
    & $\mathbf{5.63 \pm 3.29}\%$
    & $\mathbf{-47.35 \pm 5.49}\%$
    & $\mathbf{0.58 \pm 0.02}$
    & $\mathbf{8.67 \pm 0.58}$
    & $\mathbf{18.57 \pm 3.46}$ \\

    \midrule

    (d) & Heuristic
    & $69.06 \pm 2.45$
    & $29.20 \pm 2.71$
    & $42.23 \pm 2.70\%$
    & $1.93 \pm 3.62\%$
    & $-56.90 \pm 4.00\%$
    & $0.49 \pm 0.01$
    & $30.00 \pm 8.54$
    & $18.71 \pm 3.69$ \\

        & RL
    & $\mathbf{64.88 \pm 4.39}$
    & $\mathbf{20.49 \pm 3.59}$
    & $\mathbf{31.49 \pm 4.06}\%$
    & $\mathbf{-4.24 \pm 6.48}\%$
    & $\mathbf{-69.76 \pm 5.30}\%$
    & $\mathbf{0.50 \pm 0.01}$
    & $\mathbf{5.67 \pm 1.53}$
    & $\mathbf{16.65 \pm 4.70}$ \\

    \bottomrule
  \end{tabular}%
  }
\end{table*}

To justify the use of the deep reinforcement learning approach, we designed a simpler heuristic baseline and evaluated it in human-in-the-loop simulations under cases (c) and (d) in Fig. \ref{sim_setting}, using the same metrics as the previous section.

\textbf{Heuristic Baseline}: To ensure timely responsiveness to human interaction, the replanning trigger policy adopted a reactive strategy: whenever the human intervened, the robot desired path was replanned at each high-level decision interval. For the geometric parameters of the intention domain, values were determined to maximize the obstacle-free space. Specifically, the height $h$ was first extended to the maximum distance to the nearest obstacle or the boundary of the egocentric map. The radius $r$ was then incrementally expanded until both lateral edges of the cone intersected with either an obstacle or the egocentric map boundary. This method maximized the intention domain to encompass as many potential human intentions as possible given the environmental priors.

\textcolor{second}{For each case under each control condition, three independent runs were conducted, and the mean and standard deviation of each metric are reported in Table~\ref{tab:heuristic}.} The RL method achieved shorter completion time and interaction time compared to the heuristic baseline. Relative to direct human control, the RL method resulted in a smaller increase in total task completion time while achieving a greater reduction in interaction time. In the more challenging case (d), the RL method even achieved a lower task completion time than direct human control. The heuristic method produced a significantly higher number of replanning events than the RL approach, indicating that it frequently generated paths inconsistent with the human's short-term intention despite its prompt response. In addition, the RL method demonstrated superior performance in both trajectory safety and computational efficiency. Overall, these results demonstrated the advantages of jointly addressing intention domain prediction and path replanning through RL.

In the experiments, we observed that the heuristic method increased the probability of generating intention-consistent paths by performing more frequent predictions. Although this strategy achieved performance comparable to the RL method in environments with relatively sparse obstacle distributions (e.g., circular obstacle map), it struggled in scenarios involving narrow passages. Due to noise in human inputs, the actual motion direction often deviated from the corridor axis. As a result, the intention domain generated by the heuristic method could not extend across obstacles, leading the first segments of replanned paths to move toward obstacles rather than along the corridor direction. Furthermore, the high replanning frequency increased the likelihood of false triggers. Even when the human fine-tuned the robot's motion, the heuristic method could overwrite paths that were already consistent with the human intention, thereby increasing the human workload required to correct the robot's desired path. In contrast, by learning from diverse human trajectories during training, the RL approach implicitly maps environmental observations to short-term human intent. Using this representation, the policy adaptively adjusts the geometric parameters of the intention domain, generating intent-aligned paths with fewer replanning events.

To study whether our approach has a practical impact on usability and efficiency for real users, we conducted a user study using a real mobile robot with human subjects.

\section{User Study on Real Robot} 
We conducted a within-subject study on a ``search-and-rescue" task. \textcolor{modify}{In total, 10 participants (2 female, 8 male; age range 21-25 years, mean age 23.3) were recruited from a local community. \textcolor{second}{The study
protocol was reviewed and approved by the institutional research
ethics board.} All participants gave their signed consent for the experiment procedure. }

\begin{figure}
    \centering
    \includegraphics[width=8.5cm]{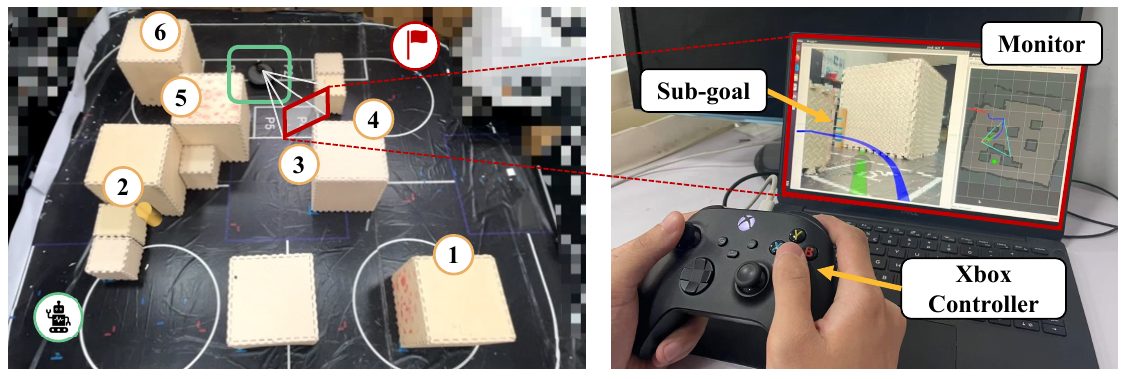}
    \caption{Physical experimental setup. Left: Experimental environment with the robot performing a search-and-rescue task. Right: The operator remotely controlled the TurtleBot 4 using an Xbox controller while monitoring the environment.}
    \label{phy_setting}
\end{figure}

Fig. \ref{phy_setting} illustrates the experimental setup. \textcolor{modify}{The experiments were conducted in an indoor area of approximately $5 \text{m} \times 5 \text{m}$. To simulate a ``search-and-rescue" scenario, six candidate sub-goal locations were arranged around foam-box obstacles. At the start of each trial, three of these locations were selected, and blue-striped cylindrical markers were placed near the obstacle edges to indicate sub-goals. The task required participants to locate and mark three sub-goals before reaching the final goal. Participants operated a TurtleBot 4 Lite robot using an Xbox controller. The user interface was identical to that used in the simulation: the left panel displayed the front camera view, while the right panel showed the occupancy grid map, the robot's desired trajectory, and task progress.} The control algorithm was implemented in ROS 2 and ran on a workstation equipped with an Intel i7-10710U CPU (1.10 GHz) and 7.7 GiB RAM, communicating with the robot via Wi-Fi. The Nav2 framework  \cite{macenski2020marathon2} was employed to handle perception, localization, planning, and control, while organizing higher-level shared control behaviors within the robot system. During model training, the regularization reward parameter $\beta = 0.5$ was used, with all other settings identical to the simulation.

\begin{figure*}[t]
    \centering
    \includegraphics[width=17.5cm]{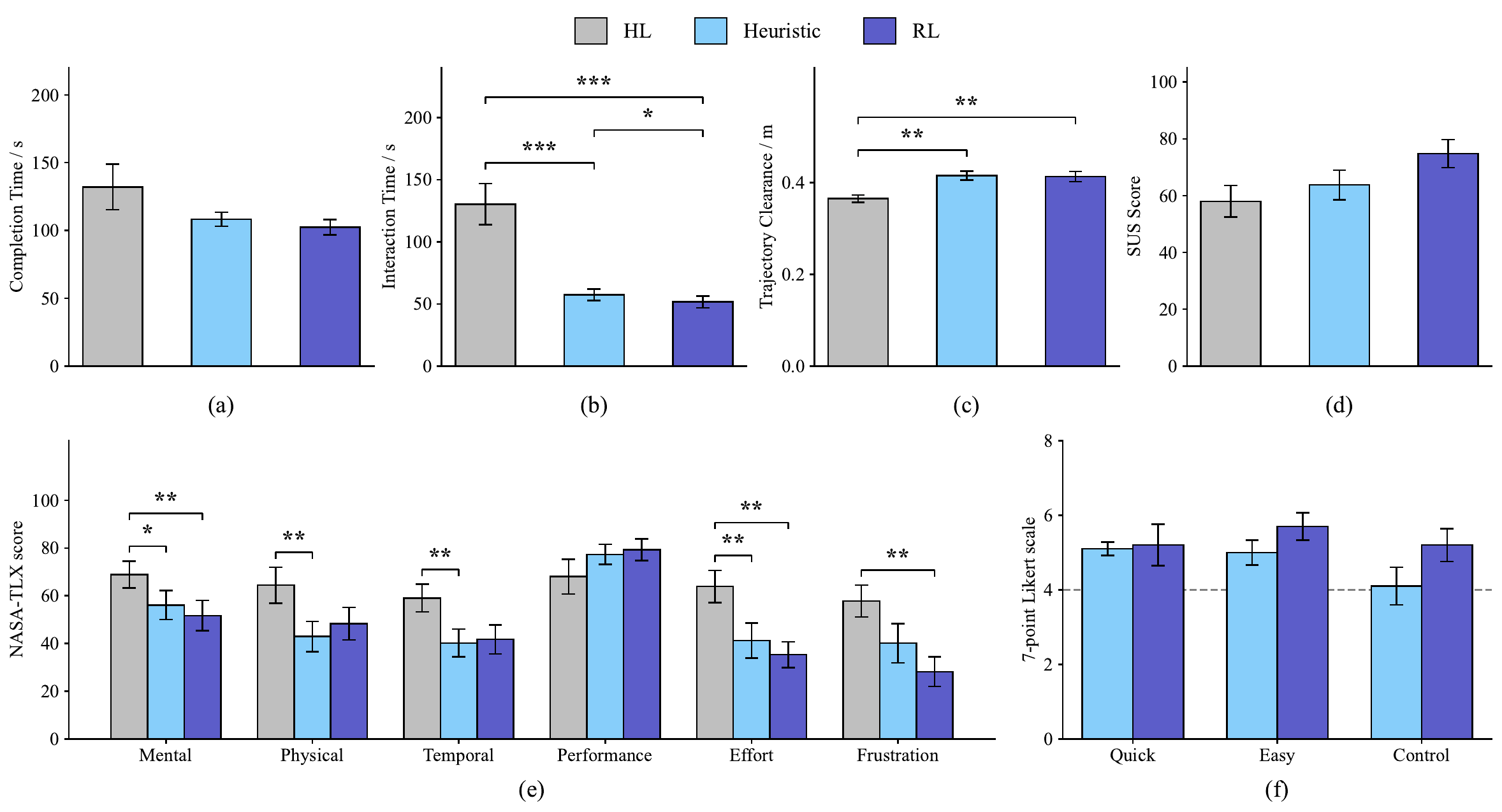}
    \caption{\textcolor{modify}{\textcolor{second}{Results of the user study. Bar plots show the mean and standard error of the evaluated measures. Asterisks indicate statistical significance (* $p < 0.05$, ** $p < 0.01$, *** $p < 0.001$). (a) Task completion time. (b) Interaction time. (c) Trajectory clearance (safety). (d) System Usability Scale (SUS) scores. (e) NASA Task Load Index (NASA-TLX) subscale scores. (f) User ratings for the task-specific statements on a 7-point Likert scale.}}}
    \label{user}
\end{figure*}

\textcolor{second}{The proposed shared-control method, which employs a learned path-replanning policy (RL), was compared with two baseline methods introduced in Section~\ref{sim_section}: direct human control (HL) and a heuristic-based shared-control method (Heuristic). The HL condition was always conducted first, enabling participants to subsequently evaluate the assistance provided by the two shared-control methods. The order of the two shared-control conditions was randomized across participants to mitigate potential order effects.}

\textcolor{modify}{Before the trials, participants were provided with a user sheet containing instructions, task descriptions, and controller mappings. To help participants better understand the robot's behavior during shared control, the sheet also illustrated how the robot replanned its desired path based on the predicted intention domain. As shown in Fig. \ref{phy_setting}, the inferred intention domain at each replanning event was highlighted as a blue triangle. They were then guided to the experimental area to familiarize themselves with the obstacle layout and the potential sub-goal locations. After that, they were given sufficient time to practice manual teleoperation under the same settings as the subsequent trials, but with randomly selected sub-goal locations.}

During the formal experiment, each participant completed one practice trial and three recorded trials under each control condition. The sub-goal distributions differed across trials but were identical between control conditions. The order of sub-goal configurations within each control condition was randomized to counterbalance potential learning effects.

{\color{modify}
After each control condition, participants evaluated their subjective workload and perceived usability of the control method. Human workload was assessed using the NASA Task Load Index (NASA-TLX) \cite{hartDevelopmentNASATLXTask1988}, a standard tool for evaluating subjective workload. Perceived usability was measured using the System Usability Scale (SUS) \cite{brookeSUSquickdirty}, an industry-standard ten-item questionnaire. \textcolor{second}{Under each shared-control condition, participants additionally rated the following statements on a seven-point Likert scale to assess the effectiveness of the robot assistance.}

\begin{itemize}

\item [1)] ``This algorithm helped me complete the task quickly."

\item [2)] ``This algorithm made it easy to complete the task."

\item [3)] ``I felt in control while using this control method."

\end{itemize}
}

\textcolor{second}{We also evaluated the performance of the three control conditions using objective measures mentioned in Section \ref{sec_sim}.}


\textcolor{second}{For each participant, the mean value of each performance metric was calculated across the three recorded trials under each control condition. For objective metrics, we employed a one-way repeated-measures ANOVA with three control conditions and ten participants with a significance threshold of $\alpha = 0.05$ to test the null hypothesis that there was no difference among the three control conditions. If the sphericity assumption was violated, Greenhouse-Geisser correction was applied. When a significant main effect was observed, post-hoc pairwise comparisons were performed using paired t-tests with Bonferroni correction. For NASA-TLX and SUS scores, Friedman's test was used, followed by Wilcoxon signed-rank tests with Bonferroni correction when a significant main effect was observed. The additional seven-point Likert-scale ratings were analyzed using Wilcoxon signed-rank tests.}

\textcolor{second}{As shown in Fig.~\ref{user} (a), no significant difference ($F(1.056, 9.506) = 4.473, p=0.061$) was observed among the three control conditions in terms of task completion time, and RL ($102.42 \pm 5.67~\mathrm{s}$) achieved a lower mean completion time than Heuristic ($108.22 \pm 5.15~\mathrm{s}$) and HL ($132.11 \pm 16.84~\mathrm{s}$). As shown in Fig.~\ref{user} (b), both Heuristic ($57.37 \pm 4.65~\mathrm{s}, t(9) = 5.637, p < 0.001$) and RL ($51.59 \pm 4.68~\mathrm{s}, t(9) = 6.319, p < 0.001$) significantly reduced interaction time compared with HL ($130.30 \pm 16.44~\mathrm{s}$). Moreover, RL further significantly reduced interaction time compared with Heuristic ($t(9) = 3.184, p = 0.033$). \textcolor{third}{These results indicate that the proposed shared control framework provides effective planning-level assistance, substantially reducing the amount of human intervention required without compromising task efficiency.}}


\textcolor{second}{As shown in Fig.~\ref{user} (c), both Heuristic ($t(9) = -4.998$, $p = 0.002$) and RL ($t(9) = -5.062$, $p = 0.002$) achieved significantly greater trajectory clearance than HL. This result suggests that the shared control methods improve system safety in cluttered environments by generating trajectories that maintain larger distances from obstacles.}


\textcolor{second}{The subjective evaluation results are summarized in Fig.~\ref{user} (d)--(f). As shown in Fig.~\ref{user} (d), $\mathrm{RL}$ achieved the highest SUS score among the three conditions, while $\mathrm{Heuristic}$ showed a moderate improvement over HL. Although the difference was not statistically significant after Bonferroni correction ($p = 0.073$), the observed trend suggests that participants perceived $\mathrm{RL}$ as the most usable control method.}


\textcolor{second}{For NASA-TLX, lower scores indicate a lower perceived workload, except for the performance subscale, for which a higher score indicates better perceived task performance. \textcolor{third}{As shown in Fig.~10 (e), significant differences were found for mental, physical, and temporal demand, effort, and frustration. Compared with HL, RL significantly reduced mental demand ($p=0.009$), effort ($p=0.005$), and frustration ($p=0.005$), while Heuristic significantly reduced mental ($p=0.012$), physical ($p=0.007$), and temporal demand ($p=0.007$), as well as effort ($p=0.005$). No significant pairwise differences were found between Heuristic and RL, and the performance subscale showed no significant overall difference ($p=0.058$).} These results further confirm that the proposed method reduces the operator's subjective workload while improving perceived task performance.}


\textcolor{second}{Finally, Fig.~\ref{user} (f) reports participants' responses to three additional seven-point Likert-scale statements evaluating the assistance provided by the two shared-control methods. \textcolor{third}{The mean ratings for both methods were above the neutral midpoint for all three statements. Although RL obtained higher mean scores than Heuristic for the Easy'' and Control'' statements, none of the three comparisons reached statistical significance (Quick: $p=0.856$; Easy: $p=0.107$; Control: $p=0.064$).} These results suggest that participants generally perceived the proposed method as easier to use and as providing a stronger sense of control.}

\textcolor{modify}{\textbf{Remark 2:} Our method works best when the operator's recent motion is consistent with a stable intended direction, especially in cluttered environments with obstacles of comparable scale. In such cases, the historical path buffer can provide reliable cues for inferring the direction of the intention domain, while the local egocentric map provides sufficient information to predict an intention domain. Nevertheless, several limitations remain. First, when the operator performs abrupt intent switches, backing-up motions, or ``peek-then-commit'' exploration in open areas, the recent motion history may not indicate a committed path change, and the predicted cone may lead to unnecessary replanning. This issue can be mitigated by adopting a more conservative replanning policy trained with an increased regularization weight $\beta$.  Second, when the environment contains obstacles with significantly different scales, the fixed local observation map may limit the size of the predicted intention domain and require multiple replanning steps to bypass large obstacles. A possible improvement is to adaptively tune the local map size according to the surrounding obstacle distribution. Third, when passing through narrow corridors, because the current historical path buffer does not encode the robot heading explicitly, the robot may need to move a short distance toward the corridor before the historical path buffer captures the operator's intention. This may still require precise human control. This limitation can be mitigated by enhancing the local shared controller with additional motion guidance based on the local environment.}

\section{Conclusion}
In this paper, we propose a novel shared control framework for mobile robots. To reduce the operator's control burden, we extend shared control to the planning level, enabling the robot to infer the operator's motion intention in real time and adjust its desired trajectory accordingly. To achieve intention-aware and safe trajectory adjustment, we propose a path replanning algorithm based on intention domain prediction. We formulate intention domain prediction and path replanning jointly as a Markov Decision Process and solve it using reinforcement learning. Both simulation and real-world experiments were conducted. In simulation, we analyzed the characteristics of the learned replanning strategy and verified its effectiveness in reducing operator workload and improving system safety. \textcolor{second}{In the real-world experiment, we conducted a user study. The statistical analysis showed that our method significantly reduced operator workload and improved safety without compromising task efficiency.}

{\color{modify}
\appendices

\section{Hyper-parameters and Network Architectures}
\label{app:network}

The hybrid actor-critic architecture used in our problem consists of two parallel actor networks and a critic network. The two actor networks are responsible for discrete action selection and continuous parameter selection, respectively. The discrete action is modeled by a categorical distribution, whereas the continuous action is modeled by a Gaussian distribution with diagonal covariance. The log standard deviation is initialized to a constant value, which determines the initial exploration scale of the policy. The critic network works as an estimator of the state-value function to avoid the overparameterization problem in parameterized action space. 

Both the policy and value functions are approximated using three-layer fully connected neural networks with 512, 256, and 128 neurons, respectively. Additional training hyperparameters are listed in Table \ref{tab:hyperparameters}.
}{\color{modify}
\begin{table}[h]
  \centering
  \caption{Hyperparameters for H-PPO Training}
  \label{tab:hyperparameters}
  \begin{tabularx}{\columnwidth}{@{} l >{\centering\arraybackslash}X @{}}
    \toprule
    Hyperparameter & Value \\
    \midrule
    Discount factor & 0.99 \\
    Maximum time steps & $2 \times 10^7$ \\
    Batch size & 512 \\
    Update epochs & 8 \\
    Learning rate & $5 \times 10^{-6}$ \\
    Clipping parameter $\epsilon$ & 0.3 \\
    Value loss coefficient & 1.0 \\
    Entropy loss coefficient & 0.01 \\
    Initial log standard deviation & -2.0 \\
    Activation function & ReLU \\
    \bottomrule
  \end{tabularx}
\end{table}
}


\bibliographystyle{Transactions-Bibliography/IEEEtran}
\bibliography{Transactions-Bibliography/IEEEabrv,Transactions-Bibliography/BIB_xx-TIE-xxxx} 

\end{document}